%% file: main.tex
\documentclass[runningheads]{llncs}
\usepackage{graphicx}
\usepackage{xcolor}
\usepackage{csquotes}
\usepackage{listings}
\usepackage[breaklinks=true,colorlinks,citecolor=royalblue,bookmarks=false]{hyperref}
\usepackage{pifont}
\usepackage{amssymb}
\usepackage{booktabs}
\usepackage{caption}
\usepackage{subcaption}

\usepackage{algorithmic}
\usepackage[ruled, vlined, linesnumbered]{algorithm2e}

\usepackage[listings,theorems]{tcolorbox}

\usepackage{tikz}
\usetikzlibrary{fit,calc}
\newcommand*{\tikzmk}[1]{\tikz[remember picture,overlay,] \node (#1) {};\ignorespaces}
\newcommand{\boxit}[2]{\tikz[remember picture,overlay]{\node[yshift=3pt,fill=#1,opacity=.25,fit={(A)($(B)+(\dimexpr\algowidth-#2\algoskipindent,.8\baselineskip)$)}] {};}\ignorespaces}

\usepackage{textcomp}

\usetikzlibrary{positioning}

\newcommand{\prioritized}[1][olive]%
    {%
    \begin{tikzpicture}[font=\sffamily]
        \node[fill=#1,minimum size=1mm,] (rect) at (0,0){};
    \end{tikzpicture}
    }

\usepackage{array}
\newcolumntype{C}[1]{>{\centering\let\newline\\\arraybackslash\hspace{0pt}}m{#1}}

\usepackage{textcomp}
\usepackage{float}
\usepackage{stfloats}
\usepackage{url}
\usepackage{cite}
\usepackage{verbatim}
\usepackage{graphicx}
\usepackage{amsmath}
\usepackage{xcolor}
\usepackage{hyperref}
\usepackage{multirow}
\usepackage{diagbox} 
\usepackage{makecell}
\usepackage{wrapfig}
\usepackage{marvosym}

\lstdefinestyle{mystyle}{
    commentstyle=\color{gray},
}

\definecolor{lighgreenbackground}{RGB}{200, 255, 200}
\definecolor{lighpurplebackground}{RGB}{241, 180, 241}

\definecolor{royalblue}{RGB}{65,105,225}
\definecolor{reference}{RGB}{128,128,128}
\definecolor{codegreen}{rgb}{0,0.6,0}
\definecolor{codegray}{rgb}{0.5,0.5,0.5}
\definecolor{codepurple}{rgb}{0.58,0,0.82}
\definecolor{backcolour}{rgb}{0.95,0.95,0.92}

\lstdefinestyle{mystyle}{
    backgroundcolor=\color{backcolour},   
    commentstyle=\color{codegreen},
    keywordstyle=\color{magenta},
    numberstyle=\tiny\color{codegray},
    stringstyle=\color{codepurple},
    basicstyle=\ttfamily\footnotesize,
    breakatwhitespace=false,         
    breaklines=true,                 
    captionpos=b,                    
    keepspaces=true,                 
    numbers=none,                    
    numbersep=5pt,                  
    showspaces=false,                
    showstringspaces=false,
    showtabs=false,                  
    tabsize=2,
}

\newcommand{\xl}[1]{{\color{orange}{[xl:#1]}}}

\def\baseline{($1+1$)-EPS}

\begin{document}
%
\title{Understanding the Importance of Evolutionary Search in Automated Heuristic Design with Large Language Models}
\titlerunning{Understanding the Importance of Evolutionary Search in AHD with LLMs}
%
\author{Rui Zhang\inst{1}\textsuperscript{(\Letter)} \and
Fei Liu\inst{1} \and
Xi Lin\inst{1} \and
Zhenkun Wang\inst{2} \and \\
Zhichao Lu\inst{1}\textsuperscript{(\Letter)} \and
Qingfu Zhang\inst{1}\textsuperscript{(\Letter)}
}
\authorrunning{R. Zhang et al.}
%
\institute{Department of Computer Science, City University of Hong Kong 
\and
School of System Design and
Intelligent Manufacturing, Southern University of Science and Technology \\
\email{rzhang.cs@gmail.com, luzhichaocn@gmail.com, qingfu.zhang@cityu.edu.hk}
}
\maketitle              
\input{0-abstract}
\input{1-introduction}

\input{2-background}
\input{3-setup}

\input{4-experiment}
\input{5-conclusion}

\begin{credits}
\subsubsection*{\ackname} The work described in this paper was supported by the Research Grants Council of the Hong Kong Special Administrative Region, China (GRF Project No. CityU11215622), the National Natural Science Foundation of China (Grant No. 62106096), the Natural Science Foundation of Guangdong Province (Grant No. 2024A1515011759), the National Natural Science Foundation of Shenzhen (Grant No. JCYJ20220530113013031).

\end{credits}

\input{6-appendix}
\let\OLDthebibliography\thebibliography
\renewcommand\thebibliography[1]{
  \OLDthebibliography{#1}
  \setlength{\parskip}{0pt}
  \setlength{\itemsep}{0pt plus 0.3ex}
}

\bibliographystyle{ieeetr}
{\bibliography{egbib}}

\end{document}

%% file: 0-abstract.tex
\begin{abstract}
Automated heuristic design (AHD) has gained considerable attention for its potential to automate the development of effective heuristics. 
The recent advent of large language models (LLMs) has paved a new avenue for AHD, with initial efforts focusing on framing AHD as an evolutionary program search (EPS) problem. 
However, inconsistent benchmark settings, inadequate baselines, and a lack of detailed component analysis have left the necessity of integrating LLMs with search strategies and the true progress achieved by existing LLM-based EPS methods to be inadequately justified. 
This work seeks to fulfill these research queries by conducting a large-scale benchmark comprising four LLM-based EPS methods and four AHD problems across nine LLMs and five independent runs. 
Our extensive experiments yield meaningful insights, providing empirical grounding for the importance of evolutionary search in LLM-based AHD approaches, 
while also contributing to the advancement of future EPS algorithmic development. 
To foster accessibility and reproducibility, we have fully open-sourced our benchmark and corresponding results.
\keywords{Automated heuristic design \and evolutionary program search \and large language model \and evolutionary computation.}
\end{abstract}

%% file: 1-introduction.tex
\section{Introduction\label{sec:intro}}

%
Automated heuristic design (AHD) aims to automatically select, refine, or construct effective heuristics, thereby obviating the necessity for rich domain expertise traditionally required in manual heuristic design~\cite{burke2013hyper, stutzle2019automated, wu2016evolutionary}. 
Considerable effort has been dedicated to employing machine learning techniques for AHD~\cite{chen2022learning, cowling2001hyperheuristic, mockus1994application}.
Among them, genetic programming (GP)~\cite{koza1994genetic} is one of the most widely used techniques for handling AHD tasks, owing to its flexible representation and efficacy across various domains \cite{langdon2013foundations, zhang2023survey, zhang2022importance, wu2016evolutionary}.
However, GP necessitates specifying a set of permissible primitives and mutation operations, which unfortunately are non-trivial and problem-dependent \cite{o2010open}. 

Recently, the advent of large language models (LLMs) has introduced novel tools for AHD. 
Preliminary endeavors have been made to model AHD as a program search problem, employing LLMs to aid the solution (i.e., heuristics) generation and optimization process within an evolutionary framework. 
For instance, FunSearch \cite{romera2024mathematical} evolves heuristics for mathematical problems that outperform existing solutions on cap set and admissible set problems \cite{tao2006additive}. 
EoH \cite{liu2024evolution} and ReEvo \cite{ye2024reevo} evolve heuristics for combinatorial optimization (CO) problems and consequently outperform existing AHD methods on traveling salesman problems (TSPs) \cite{matai2010traveling} and online bin packing (OBP) problems \cite{seiden2002online}. 

\begin{figure}[t]
\centering
\includegraphics[width=0.98\linewidth]{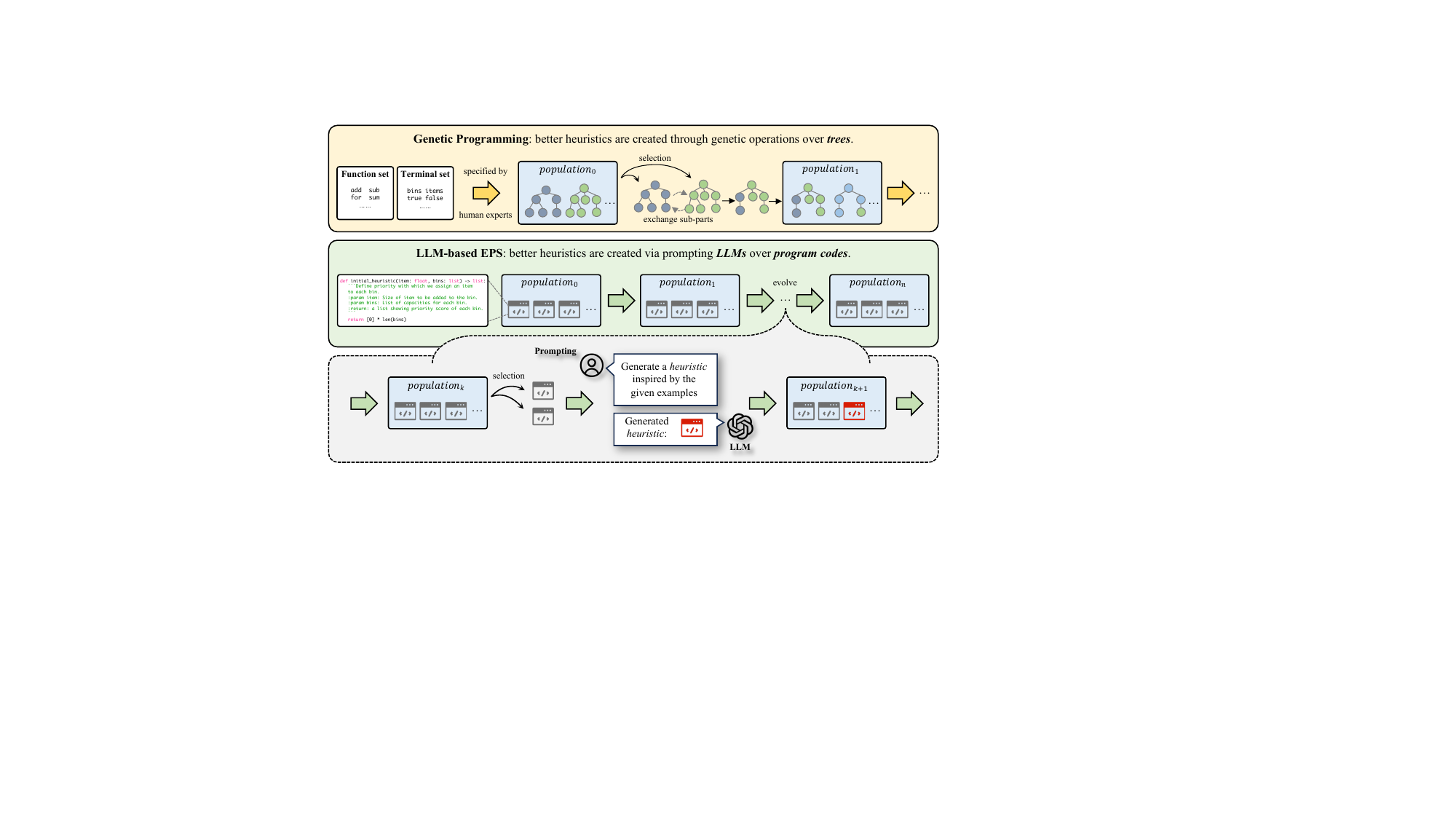}
\caption{An illustration of the LLM-based EPS paradigm, with respect to the GP-based paradigm (top section), for automated heuristic design.}
\label{fig:eps}
\vspace{-2em}
\end{figure}

These methodologies essentially adopt a canonical paradigm, referred to as \emph{LLM-based \underline{E}volutionary \underline{P}rogram \underline{S}earch (EPS)} in this work, that comprises the following key aspects: 
(\emph{i}) candidate solutions (i.e., heuristics) are represented as executable computer \emph{programs}, also referred to as codes; 
(\emph{ii}) an evolutionary computation (EC) paradigm is used to evolve toward better programs; 
and (\emph{iii}) LLMs are used as the main engine for driving the search, i.e., creating new programs, introducing variations to existing programs, etc. 
A pictorial illustration of this paradigm is provided in Fig.~\ref{fig:eps}. 

Despite a steady stream of promising empirical results, we have noticed three issues: 
(\emph{i}) Inconsistent benchmark settings, where existing LLM-based EPS methods exhibit variations in initialization, termination criteria, and choice of LLMs;  
(\emph{ii}) Inadequate baselines, where existing LLM-based EPS methods were primarily evaluated against random search or simple heuristics derived through human intuitions; 
(\emph{iii}) Lack of detailed analysis on the relative contribution of each component (e.g., choice of LLMs, prompt and search strategies, etc.) to the overall success achieved by existing LLM-based EPS methods.

To address these issues, we first develop a simple yet effective EPS baseline, dubbed \baseline{}, taking inspiration from ($1+1$)-ES \cite{hansen2016cma} and few-shot prompting \cite{brown2020language};  
and we design a unified benchmark setup comprising four LLM-based EPS methods (three existing methods and our baseline), four AHD problems, and nine different LLMs.
Then, curated experiments are designed around the following two research queries: \ding{172} the necessity of coupling LLMs with search strategies for AHD, and \ding{173} the current progress made by existing LLM-based EPS methods on AHD.  
Detailed analyses are subsequently carried out for new knowledge and insights.  

\vspace{2pt}
\noindent\textbf{Key takeaways:} Through extensive experiments, we find that:
\begin{itemize}
    \item[$\circ$] The inherent generative capability of LLMs alone is insufficient, providing an empirical justification for coupling LLMs with a search mechanism, i.e., the LLM-based EPS paradigm, for tackling AHD problems effectively. (\S\ref{subsec:expt1})

    \item[$\circ$] The performance of existing LLM-based EPS methods varies significantly across different AHD problems and LLM choices, suggesting more diverse benchmarks and applications are needed to establish a better understanding of this emergent paradigm for AHD. (\S\ref{subsec:expt2})
\end{itemize}

\vspace{2pt}
\noindent We summarize the primary contributions of this work as below:
\begin{enumerate}
    \item \textbf{Large-scale benchmark.} This work examines all existing LLM-based EPS methods along with the proposed baseline on four AHD problems under a canonical benchmark setting. Each compared method is evaluated over nine LLM choices and five independent runs.
    
    \item \textbf{Insights and implications for future research.} 
    With extensive results, we provide the empirical grounding for the necessity of LLM-based EPS for AHD and suggestions for future EPS algorithmic development.
    
    \item \textbf{Fair and reproducible evaluation.} We open-source the implementations of all compared LLM-based EPS methods, AHD problems, and interface to both open- and closed-source LLMs at \url{https://github.com/zhichao-lu/llm-eps} to foster future development. 
\end{enumerate}

%% file: 2-background.tex
\section{Background\label{sec:background}}

\noindent\textbf{Automated Heuristic Design (AHD)\label{subsec:related_ahd}} is also known as hyper-heuristics~\cite{burke2013hyper, burke2019classification, stutzle2019automated}, 
aiming to search over a space of heuristics rather than the solutions to a specific problem directly. 
Most of the AHD approaches incorporate a learning mechanism~\cite{chen2022learning, he2021automl}, such as reinforcement learning~\cite{cowling2001hyperheuristic}, Bayesian learning~\cite{mockus1994application}, case-based reasoning~\cite{burke2006case}, and evolutionary computation methods~\cite{ross1994promising, hart1998solving, terashima2005hyper, rodriguez2007combined}. 

In particular, genetic programming (GP)~\cite{koza1994genetic} has emerged as a promising approach to automate the design of heuristics.
In essence, GP maintains a set of computer programs in the form of trees, instructions, graphs, etc., where better programs are evolved through genetic operations, such as crossover and mutation.  
GP-based AHD approaches have been applied in a number of different application domains, such as combinatorial optimization~\cite{burke2006evolving, duflo2019gp, rego2011traveling}, scheduling~\cite{zhang2023survey, wu2016evolutionary}, among other areas~\cite{ drechsler1995learning, branke2015automated}. 
Although GP-based AHD approaches have achieved promising results, they are often criticized for the need to explicitly specify the function sets and primitive sets, which are not trivial and problem-dependent~\cite{o2010open}.
A more in-depth discussion of the connection between GP and the methods studied in this work is provided in Appx.~\S \ref{sec:appx_A}.

\vspace{2pt}
\noindent\textbf{Large Language Models (LLMs)\label{subsec:related_llm}} 
typically refer to deep neural networks with billions or even trillions of model parameters, built upon the Transformer architecture~\cite{vaswani2017attention}. 
The input query to LLMs can be any sequence of texts, such as natural language, codes, mathematical expressions, etc.
As output, the LLM also provides a sequence of texts in response to the input query.

With the exponential growth in model size and training data, LLMs have improved at an impressive pace in the recent past~\cite{brown2020language, achiam2023gpt}, leading to groundbreaking performance across a wide range of tasks~\cite{zhao2023survey, tian2023chatgpt, yu2023gpt, zhang2023automl}. 
Notably, the synergy between LLMs and evolutionary computation (EC) has been successfully applied to solve various optimization problems, such as prompt optimization~\cite{zhou2022large, wang2023promptagent}, algorithm design~\cite{zelikman2023self,liu2023largeb,liu2023largea}, and neural architecture search~\cite{chen2024evoprompting}, to name a few. 
Through the lens of EC, LLMs can be viewed as an intelligent variation operator~\cite{meyerson2023language}, yielding more diverse and novel offspring compared to conventional means, such as genetic operators, differential evolution, or particle swarm~\cite{hemberg2024evolving, yang2023large}. This has in turn translated to promising results in various domains~\cite{guo2023connecting, lehman2022evolution, wu2024evolutionary}.

\begin{table}[ht]
\vspace{-2em}
\centering
\caption{Existing LLM-based EPS methods for AHD along with the baseline, \baseline{}, proposed in this work.}
\label{tab:exist_eps}
\resizebox{.9\textwidth}{!}{
\begin{tabular}{@{\hspace{2mm}}l@{\hspace{2mm}}|@{\hspace{2mm}}c|@{\hspace{2mm}}c|@{\hspace{2mm}}c}
\toprule
    Method & Prompt Strategy & LLM & Search Strategy \\
    \midrule
    FunSearch \cite{romera2024mathematical} & Few-shot prompting \cite{brown2020language} & \makecell{Codey \cite{codey2024}, \\ StarCoder \cite{li2023starcoder}} & \makecell{Island model\\with re-starts} \\ \midrule
    EoH \cite{liu2024evolution} & CoT \cite{wei2022chain} & \makecell{GPT-3.5~\cite{brown2020language}, Gemini Pro,  \\ DeepSeek~\cite{guo2024deepseek}, CodeLlama~\cite{roziere2023code}}  & GA \\
    \midrule
    ReEvo \cite{ye2024reevo} & CoT \cite{wei2022chain} + Reflection \cite{brown2020language} & $2\times$GPT-3.5~\cite{brown2020language} & GA \\
    \midrule
     \begin{tabular}[c]{@{}l@{}}\baseline{}\\(our baseline)\end{tabular} & One-shot prompting & LLMs in Table~\ref{tab:llm_used} & $(1+1)$-ES \\
    \bottomrule
\end{tabular}
}
\vspace{-1em}
\end{table}

\vspace{2pt}
\noindent\textbf{Existing LLM-based EPS Methods\label{subsec:related_eps}} 
exhibit variations mainly in the following three aspects: (i) search strategy, (ii) prompt strategy, and (iii) choice of LLMs. An overview comparison is provided in Table~\ref{tab:exist_eps}. Readers are referred to Appx.~\S \ref{sec:appx_B} for elaborated descriptions.

From the perspective of search strategy, many existing EPS methods \cite{liu2024evolution,ye2024reevo} adopt the standard genetic algorithm (GA) framework, where a population of randomly initialized heuristics is made gradually better through genetic operators (i.e., crossover and mutation) and elitist selection \cite{holland1992genetic}. 
Sophisticated modifications (to the standard GA framework) have also been tried, in particular, FunSearch introduces an island model (in the form of multiple distinct populations) with a re-start mechanism to promote diversity among individuals \cite{romera2024mathematical}. 

From the perspective of prompt strategy, FunSearch \cite{romera2024mathematical} adopts a simple strategy, i.e., few-shot prompting where the LLM outputs are conditioned on a few provided examples of heuristics \cite{brown2020language}. 
More sophisticated strategies, typically variants of the chain of thought prompting (CoT) \cite{wei2022chain}, have been adopted in the subsequent works. 
In particular, EoH \cite{liu2024evolution} leverages linguistic descriptions of the heuristics (referred to as thoughts) and develops five different prompt strategies to balance the exploration and exploitation aspects of the evolutionary search; 
ReEvo \cite{ye2024reevo} applies the reflection technique \cite{shinn2024reflexion} to verbalize trends from past high-performant individuals into prompts. 

From the choice of LLM perspective, most existing EPS methods are solely evaluated with closed-source LLMs (e.g., GPT-3.5 \cite{brown2020language} and Codey \cite{codey2024}) except FunSearch which is also evaluated with an open-source LLM, i.e., StarCoder \cite{li2023starcoder}.  
In addition, ReEvo \cite{ye2024reevo} uses two LLMs -- one for generating prompts and the other one for generating heuristics. 

%% file: 3-setup.tex
\section{Preliminaries \label{sec:prelim}}
In this section, we describe the experimental setup in terms of benchmark problems, baselines, and choices of LLMs, among other settings. 

\vspace{2pt}
\noindent\textbf{Benchmark Problems.} We consider three types of applications for AHD. 

{\ding{172} Admissible Set (AS)} \cite{tao2006additive} is a variation of the cap set problem from mathematics \cite{grochow2019new}. Formally, admissible set problems, denoted as $\mathcal{A}(n, w)$, are collections of vectors in $\{0, 1, 2\}^n$ that satisfy: (1) Each vector has the same number $w$ of non-zero elements but a unique support. (2) For any three distinct vectors there is a coordinate in which their three respective values are $\{0, 1, 2\}$, $\{0, 0, 1\}$, or $\{0, 0, 2\}$. 
The objective of the admissible set problem is to maximize the size of the set while fulfilling all the aforementioned criteria. 
In this work, we set $n = 15$ and $w = 10$, i.e., $\mathcal{A}(15, 10)$, to be consistent with prior works \cite{romera2024mathematical}. 

{\ding{173} Online Bin Packing (OBP)}.
The objective of bin packing problems is to allocate a collection of items with varying sizes into the fewest possible bins of fixed capacity of $C$. We consider the online scenario where items are packed as they arrive, in contrast to the offline scenario where all items are known beforehand. 
In this work, we consider two widely used datasets for OBP: the OR dataset \cite{beasley1990or} and the Weibull dataset \cite{castineiras2012weibull}. 
To guide various LLM-based EPS methods in designing heuristics, we use 20 instances where each comprises 250 items with sizes sampled from $[20, 100]$ for the OR dataset \cite{romera2024mathematical}; 
and we use five instances where each comprises 5K items with sizes sampled from a Weibull distribution of $f(45, 3)$ for the Weibull dataset \cite{romera2024mathematical, liu2023algorithm}. The capacity $C$ of each bin is set to 150 and 100 for OR and Weibull datasets, respectively. 

{\ding{174} Traveling Salesman Problem (TSP)}
aims to find the shortest route to visit all the given locations once and return to the starting location \cite{matai2010traveling}. 
It is considered one of the most important CO problems and a widely used test bed for heuristic design approaches. 
We use a set of 64 TSP100 instances \cite{kool2018attention} where the coordinates of locations to be visited are randomly sampled from $[0, 1]$ to guide the compared LLM-based EPS methods in designing heuristics \cite{liu2023algorithm, ye2024reevo}. 

\vspace{2pt}
\noindent\textbf{Baseline.}
An adequate baseline is essential for understanding the relative improvements made by the various methods (at least empirically). 
Existing LLM-based EPS methods were mostly compared against random search (i.e., uniform sampling) or simple heuristics\footnote{For instance, an intuitive heuristic for an OBP problem could be ``place the item in the first bin with available capacity remaining''.} based on human intuitions, yielding promising performance across diverse problems. 
However, we argue that a \emph{more adequate baseline} beyond random search and intuitive heuristics is needed for a meaningful and representative comparison. 
To this end, inspired by the (1+1)-ES \cite{hansen2016cma}, we develop a simple EPS baseline, dubbed \emph{\baseline{}}.
The pseudocode of the proposed baseline is provided in Algorithm~\ref{algo:baseline}. 
Given its simplistic design in both the search and prompt strategies, we envision that \baseline{} should simulate the lower bound of the performance of the EPS paradigm. 

\begin{minipage}{0.5\textwidth}
\begin{algorithm}[H]
\SetAlgoLined
\SetKwInOut{Input}{Input}
\SetKwInOut{Output}{Output}
\SetKwFor{For}{for}{do}{end for}
\footnotesize
\Input{$f_{\rm LLM}$: a LLM,\\$h_{\rm T}$: a template heuristic,\\$T$: max. \# of gens.}
    $h_{\rm best}$ $\leftarrow$ $h_{\rm T}$ \textcolor{gray}{// initialize the best heuristic (found so far) to $h_{\rm T}$}.\\
    $s_{\rm best}$ $\leftarrow$ $evaluate(h_{\rm best})$ \textcolor{gray}{// evaluate the performance score of $h_{\rm best}$}.\\
    \While{$t < T$}{
        \tikzmk{A}$prompts$ $\leftarrow$ $One\mbox{-}shot\mbox{ }prompting(h_{best})$ \textcolor{gray}{//create inputs to $f_{\rm LLM}$}\tikzmk{B}\boxit{blue!30}{2} \textcolor{gray}{by converting $h_{best}$ to prompts via one-shot prompt engineering}.\\
        $h$ $\leftarrow$ $f_{\rm LLM}(prompts)$ \textcolor{gray}{// create a heuristic via a LLM}.\\
        $s$ $\leftarrow$ $evaluate(h)$ \textcolor{gray}{// evaluate the performance of $h$}. \\
        \If{$s < s_{\rm best}$}{
            \textcolor{gray}{// update the best heuristic found so far and its score.}
            $h_{\rm best}$ $\leftarrow$ $h$, $s_{\rm best}$ $\leftarrow$ $s$
        }
    }
\textbf{Return} $h_{\rm best}$, $s_{\rm best}$
\caption{\baseline{} \label{algo:baseline}}
\end{algorithm}
\end{minipage}
\hfill
\begin{minipage}{0.46\textwidth}
\begin{tcolorbox}[colback=blue!5,colframe=blue!75!black,title=One-shot Prompting]
\begin{itemize}
\setlength\itemsep{5pt}
    \small
    \item[Idea:] Create input prompts ($h$) by providing the best heuristic ($h_{\rm best}$) found so far as an example.
    \item[E.g.:] The\colorbox{backcolour}{ shaded texts }below are prompts created for an online bin packing problem. 
\end{itemize}
\vspace{-1em}
\begin{lstlisting}[language=Python, escapeinside={(*}{*)}]{Name}
def (*$h_{\rm best}$*)(
  item: float, 
  bins: list) -> list:
  priority = </> # omitted for brevity
  return priority
  
def (*$h$*)(
  item: float, 
  bins: list) -> list:
  priority = <...to be filled by a LLM...>
  return priority
\end{lstlisting}
\end{tcolorbox} 
\end{minipage}

\begin{table}[t]
\centering
\caption{Overview of the LLMs evaluated in this work. We use performance on ``HumanEval'' \cite{chen2021evaluating} and ``MMLU''~\cite{hendrycks2020measuring} to indicate the capabilities of LLMs on code and general knowledge, respectively. Both metrics are greater the better. \label{tab:llm_used}}
\resizebox{.85\textwidth}{!}{%
\begin{tabular}{@{\hspace{2mm}}l|@{\hspace{2mm}}c@{\hspace{2mm}}|@{\hspace{2mm}}c@{\hspace{2mm}}c@{\hspace{2mm}}|@{\hspace{2mm}}c@{\hspace{2mm}}c@{\hspace{2mm}}}
\toprule
Model & $^{\#}$P & \begin{tabular}[c]{@{}c@{}}Specialized\\for Code\end{tabular} & \begin{tabular}[c]{@{}c@{}}Open \\ Source\end{tabular} & \begin{tabular}[c]{@{}c@{}}HumanEval ($\uparrow$) \\ \cite{chen2021evaluating} \end{tabular} & \begin{tabular}[c]{@{}c@{}} MMLU ($\uparrow$) \\ \cite{hendrycks2020measuring} \end{tabular} \\ \midrule
UniXcoder~\cite{guo2022unixcoder} & 0.3B & \ding{53} & \checkmark & - & -  \\\midrule
StarCoder~\cite{li2023starcoder} & 15.5B & \checkmark & \checkmark  & 33.6\% & - \\ \midrule
\multirow{2}{*}{CodeLlama \cite{roziere2023code}} & 7B & \checkmark  & \checkmark & 34.8\% & 42.1\% \\
 & 34B & \checkmark & \checkmark  & 48.8\% & 53.1\% \\ \midrule 
\multirow{2}{*}{DeepSeek-Coder \cite{guo2024deepseek}} & 6.7B & \checkmark & \checkmark  & 66.1\% & 34.6\% \\
 & 33B & \checkmark & \checkmark & 69.2\% & 39.5\% \\ \midrule
GPT-3.5~\cite{brown2020language} & - & \ding{53} & \ding{53} & 60.3\% & 70.0\% \\ \midrule
GPT-4~\cite{achiam2023gpt} & 1.76T & \ding{53} & \ding{53} & 76.5\% & 86.4\% \\ \midrule
Claude 3 Opus~\cite{anthropic2024Claude3} & 137B & \ding{53} & \ding{53} & 84.9\% & 86.8\% \\ \bottomrule
\end{tabular}%
}
\vspace{-1em}
\end{table}

\vspace{2pt}
\noindent\textbf{Choice of LLMs.} 
We consider a diverse set of LLMs to investigate the impact of the choice of LLMs on the AHD performance. 
We include five open-source LLMs that were fine-tuned on code-related tasks and three closed-source LLMs developed for general purposes. 
In particular, we consider the most powerful LLM currently available, i.e., Claude 3 Opus~\cite{anthropic2024Claude3}, and the most capable open-source LLM for coding tasks, i.e., DeepSeek-Coder~\cite{guo2024deepseek}. 
For completeness, we also include the most powerful coding language model prior to the LLM era, i.e., UniXcoder~\cite{guo2022unixcoder}. 
Table~\ref{tab:llm_used} provides an overview of the considered LLMs.
For open-source LLMs, we deploy these models locally on a server with 16 NVIDIA V100 GPU cards; 
while for closed-source LLMs, we rely on the respective APIs provided by OpenAI and Anthropic to get responses.

\vspace{2pt}
\noindent\textbf{Evaluation Metric.}
To evaluate the AHD performance, we report the mean relative distance (or gap) in performance, $\Delta_{\rm{d}}$, of the obtained heuristic with respect to the performance of the best-known performance, mathematically as follows. 
$$ \Delta_{\rm{d}} = 100\% \times \frac{1}{N_p}\sum_{p=1}^{N_p}\frac{1}{N_m}\sum_{m=1}^{N_m}\frac{(-1)^{I_{p}}(M_{p,m} - M_{p}^{\ast})}{M_{p}^{\ast}}$$
\noindent where $N_p$ is the number of compared problems, $N_m$ is the number of considered LLMs, $M_{p,m}$ is the performance of a heuristic for the $p$-th problem with $m$-th LLM. $M_{p}^{\ast}$ is the best-known performance for the $p$-th problem. And $I_{p}$ is one if a higher value indicates better performance for the $p$-th problem (i.e., for maximization problems) and zero otherwise (i.e., for minimization problems).

\vspace{2pt}
\noindent\textbf{Other Settings.}
We initialize all compared methods with the respective template heuristic on each problem. 
The details of the template heuristics are provided in Appx. \S \ref{sec:appx_C}. 
%
%
For EoH \cite{liu2023algorithm}, we initially incorporate a template heuristic into the initial population, and repeatedly apply crossover and mutation operators to the existing individuals in the population to generate new individuals until the desired population size is achieved.
For both EoH \cite{liu2023algorithm} and ReEvo \cite{ye2024reevo}, we increase the population size as well as the maximum number of evaluations. 
We perform ablation studies on this in Appx. \S \ref{sec:appx_expt1}.
Table~\ref{tab:bench_setting} summarizes the benchmark hyper-parameter settings. 

\begin{table}[ht]
\caption{Summary of benchmark settings.}
\centering
\resizebox{.85\textwidth}{!}{%
\label{tab:bench_setting}
\begin{tabular}{@{\hspace{2mm}}l|@{\hspace{2mm}}c}
\toprule
Description of Setting & Value \\ \midrule
Maximum number of function evaluations (\#FE) & 10,000 \\ 
Population size (for EoH and ReEvo) & 100 \\ 
\# of islands, \# of samples per prompt (for FunSearch) & 10, 4 \\
Number of independent runs per experiment & 5 \\ \midrule 
\begin{tabular}[c]{@{}l@{}}Maximum evaluation time for each heuristic\\(to cope with invalid heuristics, such as infinite loops) \end{tabular} & \begin{tabular}[c]{@{}l@{}} 50 sec (TSP);\\20 sec (others) \end{tabular}\\
\bottomrule
\end{tabular}
}
\end{table}

%% file: 4-experiment.tex
\begin{figure}[t]
     \centering
     \begin{subfigure}[b]{0.95\textwidth}
         \centering
         \includegraphics[width=\textwidth]{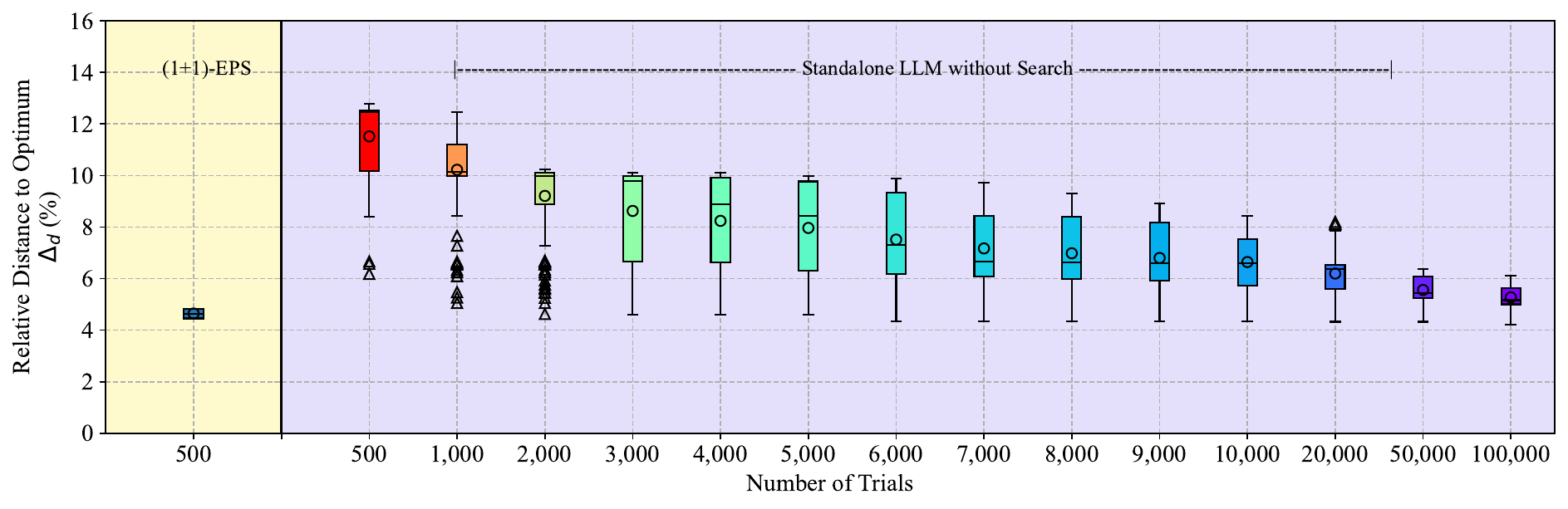}
         \caption{Top-$5$\textperthousand{} heuristics}
         \label{fig:llm_alone_top5_1000}
     \end{subfigure}\\
     \begin{subfigure}[b]{0.95\textwidth}
         \centering
         \includegraphics[width=\textwidth]{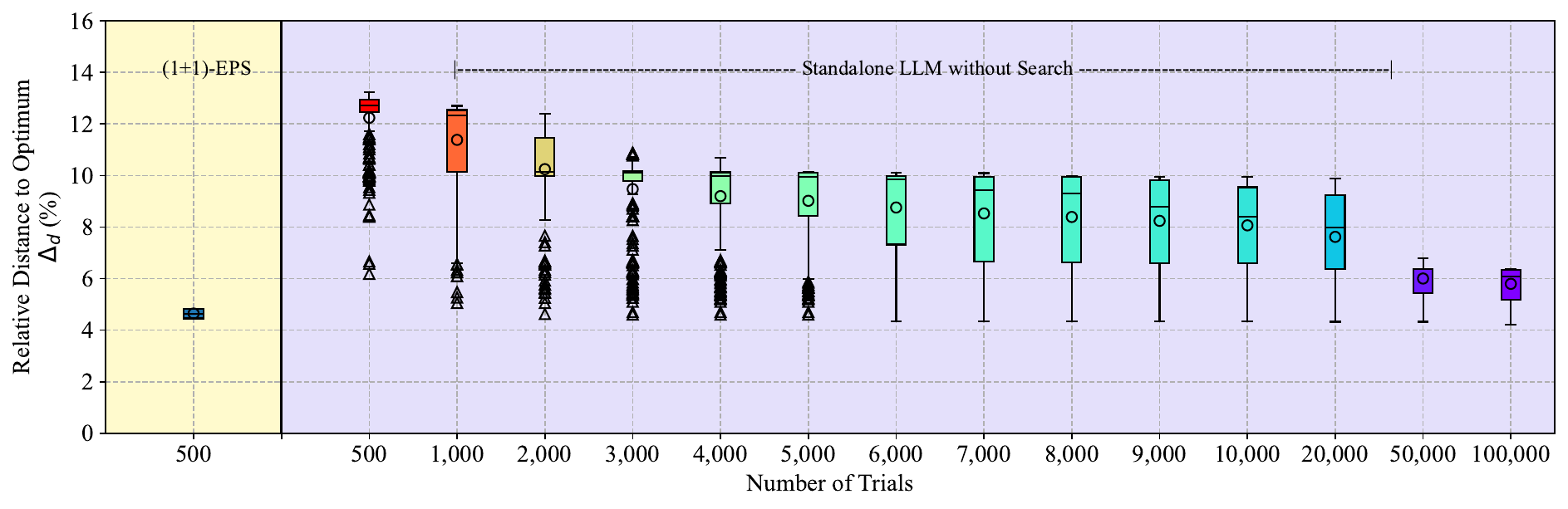}
         \caption{Top-$1\%$ heuristics}
         \label{fig:llm_alone_top1_100}
     \end{subfigure}
    \caption{Box plot comparison on the performance of the top-\{(a) $5$\textperthousand{}, (b) $1\%$\} heuristics generated by GPT-3.5 under various query budgets. The performance is measured as the relative distance to the best-known optimum ($\Delta_{\rm d}$) aggregated over four AHD problems and five independent runs. Lower $\Delta_{\rm d}$ indicates better performance. The performance of the simple baseline \baseline{} with GPT-3.5 under a small query budget of 500 is also provided as a reference.}
    \label{fig:llm_alone}
\end{figure}

\section{Experimental Results and Analyses}

\subsection{Performance of Standalone LLMs on AHD\label{subsec:expt1}}
\noindent\textbf{Motivation.} 
Standalone LLMs have consistently showcased exceptional performance across a diverse array of AI applications, reaching a point where the research community has come to expect impressive results from them on new and challenging tasks.
To this end, we wonder whether the inherent generative capability of LLMs alone (without coupling with an evolutionary search mechanism) would suffice for AHD tasks.
In this work, we attempt to answer this question from the following two angles. 

\subsubsection{Angle I: Impact of Query Budget\label{subsubsec:expt1_angle1}\\}
\mbox{}

\vspace{-5pt}
\noindent\textbf{Experimental Design.} Firstly, we aim to validate the performance of standalone LLMs on AHD problems under different query budgets, i.e., maximum \# of queries allowed to be sent to LLMs. 
Given an AHD problem, we provide the problem context along with the template heuristic (i.e., $f_{T}$ in Algorithm~\ref{algo:baseline}) as prompts to a LLM, 
and we ask it to keep generating new heuristics until query budgets are exhausted. 
Note that we do not proactively check for duplicate heuristics simply because no effective tools for functionality-level duplicate detection are readily available. 

\vspace{2pt}
\noindent\textbf{Results.} 
Fig.~\ref{fig:llm_alone} depicts the aggregated performance (i.e., mean $\Delta_{\rm d}$ over four AHD problems) of the heuristics generated by standalone GPT-3.5 with various query budgets. 
We also include the performance of the heuristics obtained by our baseline \baseline{} as a reference. 
Due to space constraints, more detailed results on individual AHD problems with different LLMs are discussed in Appx.~\S\ref{sec:appx_expt2}. 

Our analysis reveals that while the performance of standalone LLMs on AHD problems generally improves with increasing query budgets, several critical observations emerge:
\begin{itemize}
    \setlength\itemsep{2pt}
    \item[$\circ$] There remains a significant gap between the performance of heuristics generated by standalone LLMs and the best-known performance (indicated by $\Delta_{\rm d}=0$, i.e., x-axis in Fig.~\ref{fig:llm_alone}), even with a substantial query budget of 100,000.
    
    \item[$\circ$] Although there is a steady improvement in the mean performance of the top-ranked heuristics, the performance of the best individual heuristics (represented by the lower bars of the boxes) shows minimal enhancement as query budgets increase.
    
    \item[$\circ$] Standalone LLMs are highly ineffective\footnote{The ineffectiveness is in the sense that a simple EPS baseline achieves better mean performance with much lower variances than standalone LLMs with an order of magnitude more query budget, as depicted in Fig.~\ref{fig:llm_alone}.} on AHD problems, even when granted an order of magnitude more queries. 
\end{itemize}
\noindent In summary, these observations suggest that merely increasing the number of attempts by a standalone LLM is insufficient for effectively addressing AHD problems. This underscores \emph{the need to integrate LLMs with search methods to enhance their efficacy in AHD contexts}.  

\begin{figure}[t]
     \centering
     \begin{subfigure}[b]{0.95\textwidth}
         \centering
         \includegraphics[width=\textwidth]{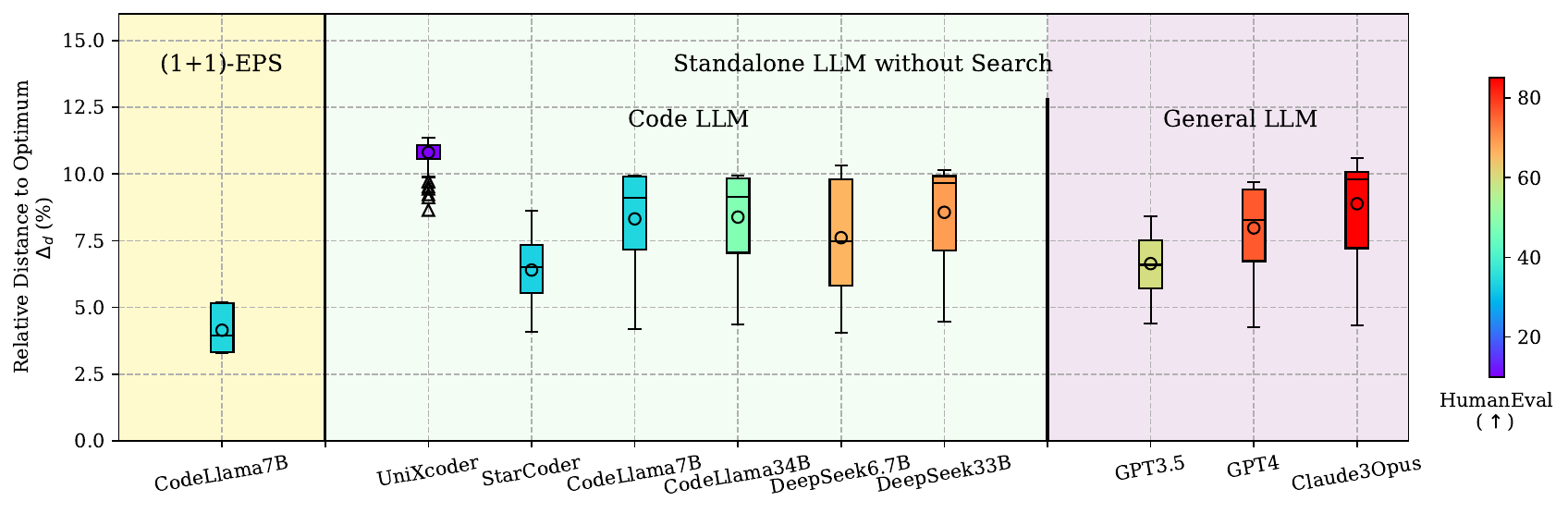}
         \label{fig:llm_alone_top_5_1000_model_choice}
     \end{subfigure}
    \caption{Box plot comparison on the performance of the top-$5$\textperthousand{} heuristics generated by LLMs with varying capacities under 10,000 query budgets. 
    We group LLMs into two categories: (1) LLMs specialized for coding tasks (with background shaded in \prioritized[lighgreenbackground]) and (2) general-purpose LLMs (with background shaded in \prioritized[lighpurplebackground]).
    Then, the LLMs are arranged in the order of ascending model size within each group. 
    The color scale of the boxes corresponds with the scores on HumanEval \cite{chen2021evaluating}.
    The performance is measured as the relative distance to the best-known optimum ($\Delta_{\rm d}$) aggregated over four AHD problems and five independent runs. Lower $\Delta_{\rm d}$ indicates better performance. 
    The performance of the simple baseline \baseline{} with CodeLlama-7B is also provided as a reference.
    }
    \label{fig:llm_alone_model_choice}
\end{figure}

\begin{table}[t]
\caption{The performance of the top-1 heuristics generated by LLMs with varying capacities. The performance is measured as the relative distance to the best-known optimum ($\Delta_{\rm d}$) aggregated over four AHD problems and five independent runs. Lower $\Delta_{\rm d}$ indicates better performance.}
\label{tab:llm_alone_model_choice_top1}
\centering
\resizebox{.98\textwidth}{!}{
    \begin{tabular}{c|c|c|c|c|c|c|c|c}
    \toprule
    UniXcoder& \makecell{DeepSeek\\-6.7B } &\makecell{CodeLlama\\-7B} & \makecell{StarCoder\\15.5B}&  \makecell{DeepSeek\\-33B} & \makecell{CodeLlama\\-34B} & GPT-3.5 & \makecell{Claude 3\\ Opus} & GPT-4 \\ \midrule
    9.15\% & 4.24\% & 4.32\% & 4.21\% & 4.59\% & 4.48\% & 4.31\% & 4.63\% & 4.44\% \\
    \bottomrule
    \end{tabular}
}
\end{table}

\subsubsection{Angle II: Impact of More Capable LLMs\label{subsubsec:expt1_angle2}\\}
\mbox{}

\vspace{-5pt}
\noindent\textbf{Experimental Design.} Next, we attempt to understand the relationship between LLMs' capacity and their performance on AHD problems. 
In this work, we consider the model size (in terms of \# of parameters), the coding performance (in terms of HumanEval scores \cite{chen2021evaluating}), and the general performance across many tasks (in terms of MMLU scores \cite{hendrycks2020measuring}) as proxy indicators for measuring a LLM's capacity.
A diverse set of nine different LLMs is considered, with more details provided in Table~\ref{tab:llm_used}.
Specifically, given an AHD problem, we provide the problem context along with the template heuristic (i.e., $f_{T}$ in Algorithm~\ref{algo:baseline}) as prompts to a LLM, 
and we ask it to keep generating new heuristics until the query budget of 10,000 is exhausted. 

\vspace{2pt}
\noindent\textbf{Results.} 
Fig.~\ref{fig:llm_alone_model_choice} depicts the aggregated performance (i.e., mean $\Delta_{\rm d}$ over four AHD problems) of the heuristics generated by LLMs with varying capacities. 
We also include the performance of the heuristics obtained by our baseline \baseline{} with CodeLlama-7B (i.e., a small-capacity LLM) as a reference. 
In addition, we provide the performance of the top-1 heuristics generated by various LLMs in Table~\ref{tab:llm_alone_model_choice_top1}. 
Constrained by space, more elaborated results on individual AHD problems with different thresholds on filtering top heuristics are provided in Appx.~\S \ref{sec:appx_expt3}. 
\interfootnotelinepenalty=10000
Evidently, we make the following observations:
\begin{itemize}
    \setlength\itemsep{2pt}
    \item[$\circ$] LLMs with more capacity (i.e., more \# of model parameters, better HumanEval and MMLU scores) do not necessarily lead to better performance on AHD problems. 
    
    \item[$\circ$] LLMs fine-tuned for coding tasks (i.e., the group with background shaded in \prioritized[lighgreenbackground]) are not statistically better than general purpose LLMs (i.e., the group with background shaded in \prioritized[lighpurplebackground]).
    
    \item[$\circ$] Standalone LLMs with large overall model capacity are still highly ineffective\footnote{The ineffectiveness is in the sense that a low-capacity LLM coupled with a simple EPS baseline significantly standalone LLMs with orders of magnitude more model capacities (e.g., GPT-4 and Claude 3 Opus), as depicted in Fig.~\ref{fig:llm_alone_model_choice}.} on AHD problems.

    \item[$\circ$] Conventional LLMs, i.e., variants of BERT \cite{devlin-etal-2019-bert} such as UniXcoder \cite{guo2022unixcoder}, are significantly inferior to modern LLMs (e.g., GPTs) on AHD problems. 
\end{itemize}
\noindent In summary, the above observations suggest that simply importing more capable LLMs is insufficient for tackling AHD problems, reinforcing \emph{the need to integrate LLMs with search methods to enhance their efficacy in AHD contexts}.

\subsubsection{Summary and Implications\\}
\mbox{}

\vspace{-5pt}
\noindent Observations from the previous sections have converged to a consensus that the inherent generative capability of \textbf{LLMs alone is insufficient for AHD} problems, which holds true under increased query budget (\S\ref{subsubsec:expt1_angle1} Angle I) and model capacity (\S\ref{subsubsec:expt1_angle2} Angle II), suggesting the \textbf{necessity of coupling LLMs with a search strategy} to tackle AHD problems effectively.

Given the modular yet flexible framework, we believe that the LLM-based EPS paradigm, \textbf{synergizing LLMs with an evolutionary search strategy}, is a meaningful approach to addressing the general AHD problems. 

\begin{figure}[t]
     \centering
     \begin{subfigure}[b]{0.49\textwidth}
         \centering
         \includegraphics[width=\textwidth]{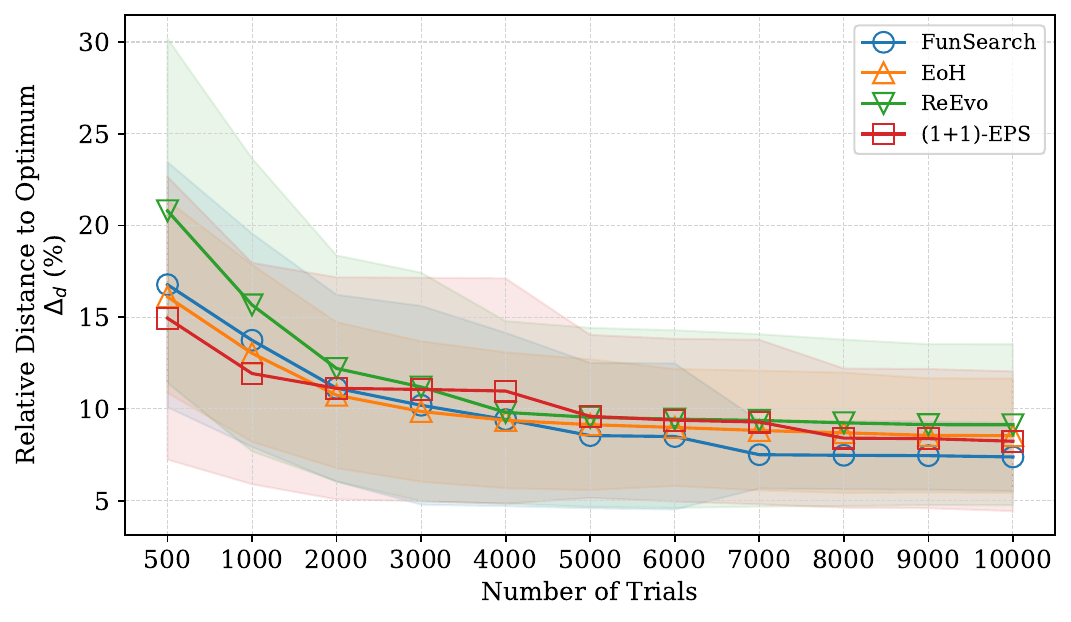}
         \caption{Admissible Set}
         \label{fig:err_bar_as}
     \end{subfigure}\hfill
     \begin{subfigure}[b]{0.50\textwidth}
         \centering
         \includegraphics[width=\textwidth]{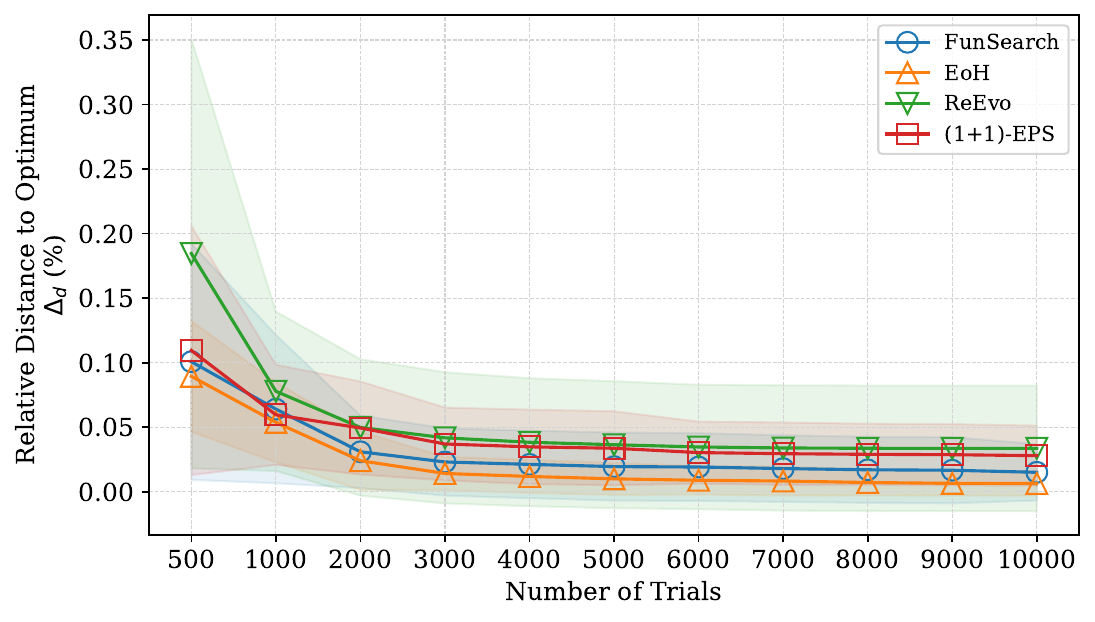}
         \caption{Travelling Salesman}
         \label{fig:err_bar_tsp}
     \end{subfigure}\hfill
     \begin{subfigure}[b]{0.49\textwidth}
         \centering
         \includegraphics[width=\textwidth]{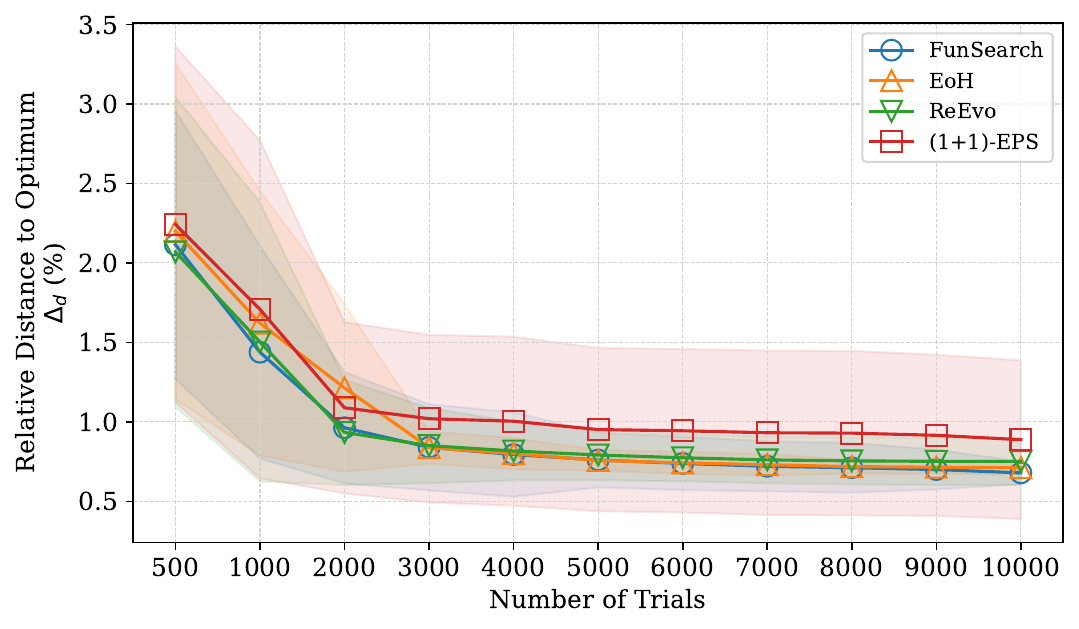}
         \caption{Online Bin Packing (Weibull)}
         \label{fig:err_bar_wei}
     \end{subfigure} \hfill
     \begin{subfigure}[b]{0.49\textwidth}
         \centering
         \includegraphics[width=\textwidth]{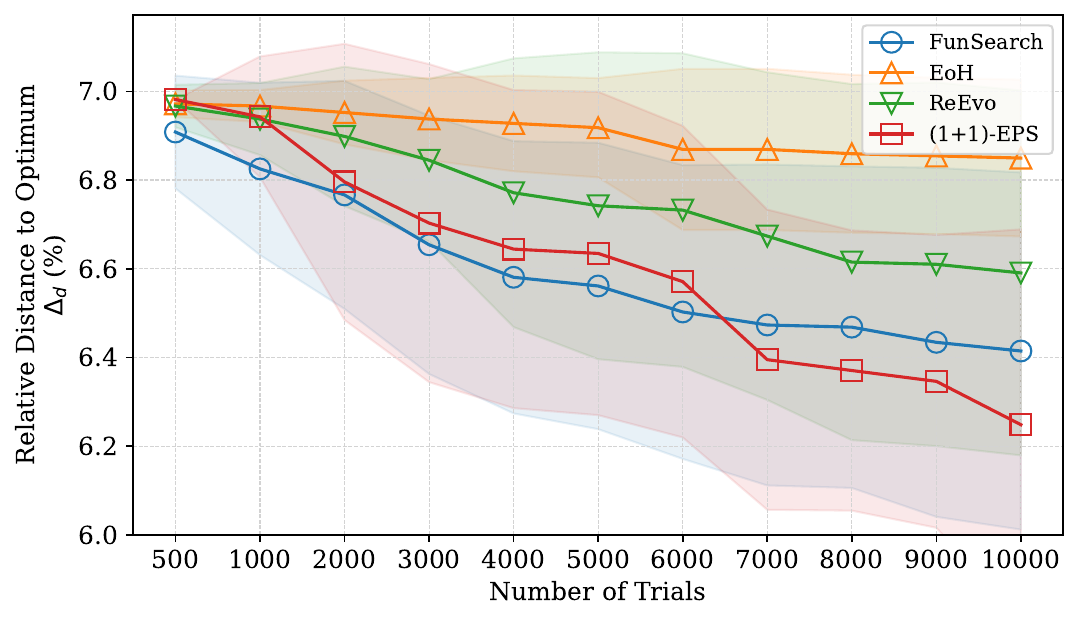}
         \caption{Online Bin Packing (OR)}
         \label{fig:err_bar_or}
     \end{subfigure}
    \caption{Convergence curve comparison on the performance of the top-1 heuristics achieved by various EPS methods. The mean relative distances to the best-known optimum ($\Delta_{\rm d}$) averaged over five independent runs are denoted with markers, while the standard deviations of $\Delta_{\rm d}$ are shown with the shaded regions.}
    \label{fig:convergence_four_task}
\end{figure}

\subsection{Performance of Existing LLM-based EPS Methods on AHD\label{subsec:expt2}}
We decompose our investigations into the following two angles to establish an empirical understanding of the progress made by the existing LLM-based EPS methods on AHD.

\subsubsection{Angle I: Relative Improvements over Adequate Baseline\label{subsubsec:expt2_angle1}\\}
\mbox{}

\noindent\textbf{Motivation.}
Existing LLM-based EPS methods incorporate a variety of complications in the search and the prompt components (see \S\ref{subsec:related_eps} for more details). 
The relative improvements contributed by these modifications are primarily evaluated against random search or simple heuristics derived through human intuitions. 
On the one hand, whether the observed improvements over these naive baselines meaningfully capture the advancement in algorithmic design remains questionable; 
while on the other hand, the general utility of the enhancements introduced by various EPS methods also remains to be further evaluated.

\vspace{2pt}
\noindent\textbf{Experimental Design.} We benchmark existing LLM-based EPS methods (i.e., FunSearch \cite{romera2024mathematical}, EoH \cite{liu2023algorithm}, and ReEvo \cite{ye2024reevo}) against the proposed baseline \baseline{} on four AHD problems with seven LLMs\footnote{We exclude UniXcoder and StarCoder from Table~\ref{tab:llm_used} as they are mainly designed for code completion, which are not compatible with EoH and ReEvo that also require comprehension of natural languages.}. 
We repeat each experiment five times with different random seeds. 
All other benchmark settings are identical to those described in \S\ref{sec:prelim} unless otherwise specified.

\vspace{2pt}
\noindent\textbf{Results.} 
Fig.~\ref{fig:convergence_four_task} compares the aggregated performance (i.e., mean $\Delta_{\rm d}$ over seven LLMs and five independent runs) among existing EPS methods and the proposed baseline on four AHD problems. 
Evidently, we observe that:
\begin{itemize}
    \setlength\itemsep{2pt}
    \item[$\circ$] Performance varies significantly across different problems for all existing LLM-based EPS methods, with no single method demonstrating consistent superiority.
    
    \item[$\circ$] Specifically, the EoH method consistently outperforms all others in the TSP problem throughout the search process, while the simple baseline \baseline{} shows competitive performance, except in the OBP (Weibull) problem. 
\end{itemize}
\noindent These empirical findings suggest that there may not be universally effective and efficient LLM-based EPS method for all AHD problems, reinforcing the applicability of the ``no free lunch'' (NFL) theorem to AHD. 

\begin{figure}[t]
     \centering
     \begin{subfigure}[b]{0.49\textwidth}
         \centering
         \includegraphics[width=\textwidth]{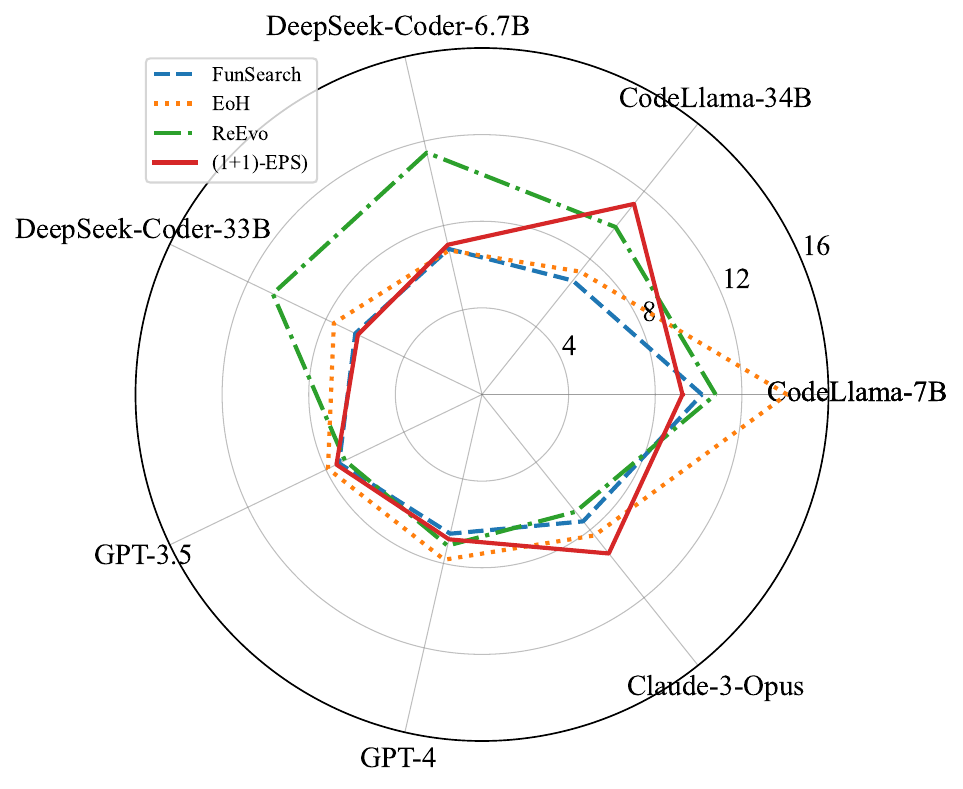}
         \caption{Admissible Set}
         \label{fig:radar_admi}
     \end{subfigure}\hfill
     \begin{subfigure}[b]{0.49\textwidth}
         \centering
         \includegraphics[width=\textwidth]{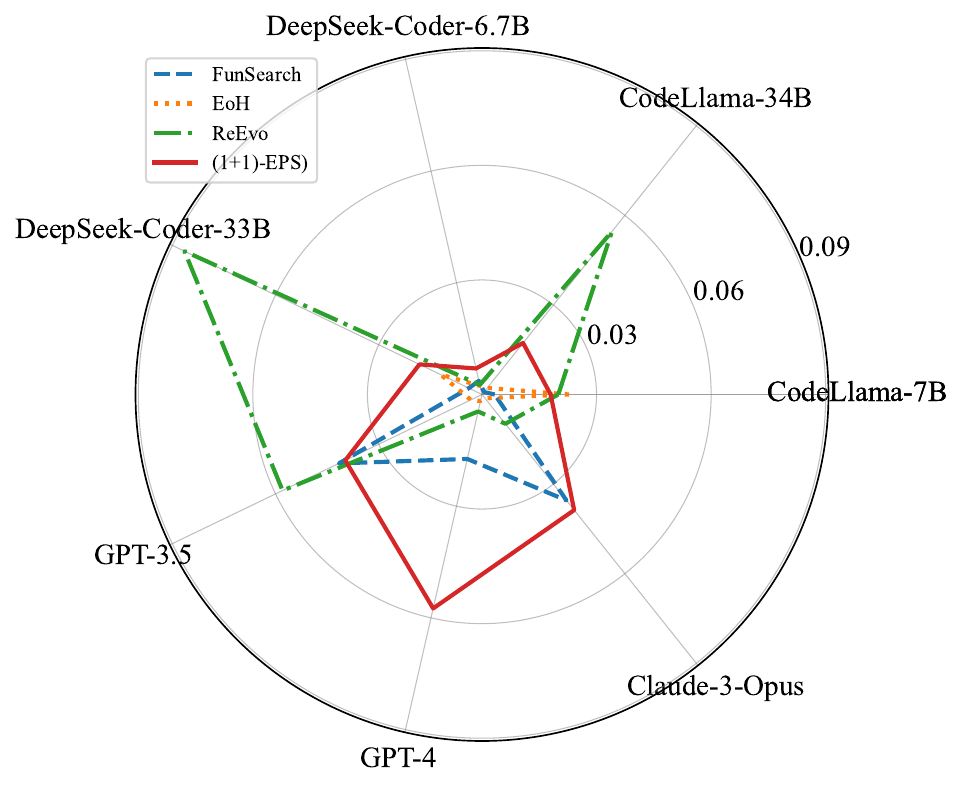}
         \caption{Travalling Salesman}
         \label{fig:radar_tsp}
     \end{subfigure} \\
     \begin{subfigure}[b]{0.49\textwidth}
         \centering
         \includegraphics[width=\textwidth]{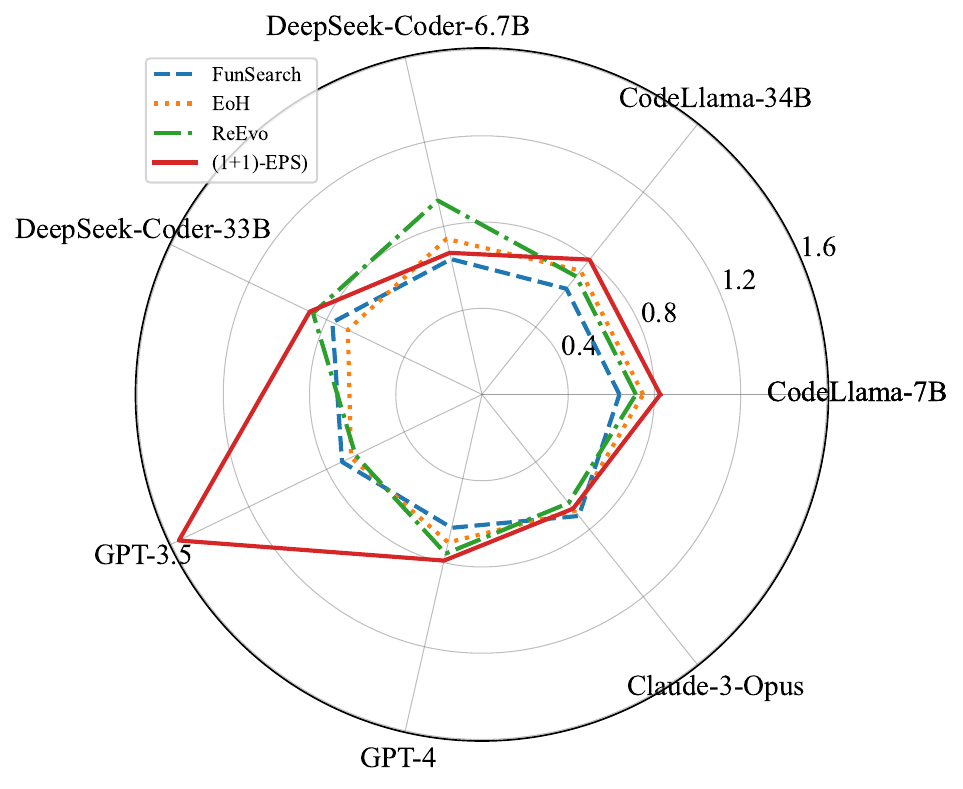}
         \caption{Online Bin Packing (Weibull)}
         \label{fig:radar_wei}
     \end{subfigure} \hfill
     \begin{subfigure}[b]{0.49\textwidth}
         \centering
         \includegraphics[width=\textwidth]{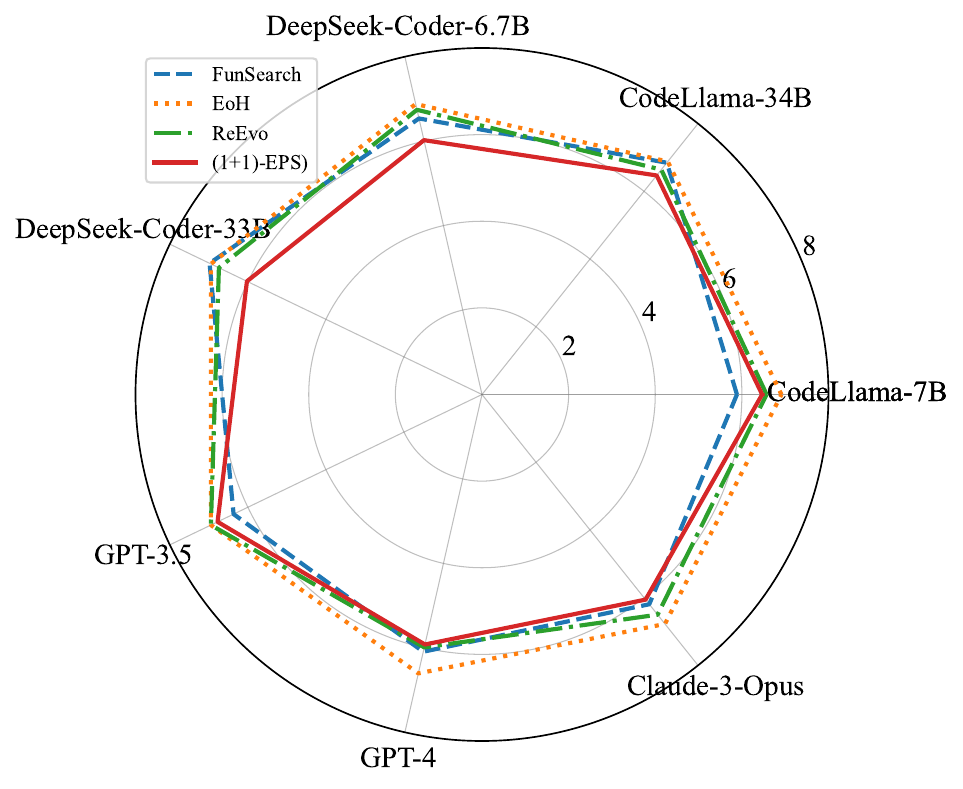}
         \caption{Online Bin Packing (OR)}
         \label{fig:radar_or}
     \end{subfigure}
    \caption{Radar plot comparison on the performance of the top-1 heuristics achieved by various EPS methods with different choices of LLMs. The radius of each vertex is calculated by the mean relative distances to the best-known optimum ($\Delta_{\rm d}$) averaged over five independent runs; hence, a smaller radius/enclosed area indicates better performance. 
    }
    \label{fig:radar_four_task}
    \vspace{-2em}
\end{figure}

\subsubsection{Angle II: Dependency on the Choice of LLMs \label{subsubsec:expt2_angle2}\\}
\mbox{}

\vspace{-5pt}
\noindent\textbf{Motivation.} 
Existing LLM-based EPS methods are typically evaluated using only one particular choice of LLMs \cite{liu2023algorithm, ye2024reevo, ma2024eureka}. 
This raises uncertainty regarding the extent to which performance enhancements suggested by these methods can be applied to other LLM choices. 
Compounding this issue, the predominant LLM utilized in these EPS methods, i.e., GPT-3.5, is closed-source in nature.
Should the efficacy of existing EPS methods hinge significantly on closed-source LLMs, the geographically restricted access to APIs may impede future development built upon these methods.

\vspace{2pt}
\noindent\textbf{Experimental Design.} The experimental setup is identical to those described in \S\ref{subsubsec:expt2_angle1}, except on the utilization of different LLMs. 
In this case, we do not aggregate experiments across various LLMs. Instead, we aim to directly compare the performance under different LLM choices for each EPS method.  

\vspace{2pt}
\noindent\textbf{Results.}
Fig.~\ref{fig:radar_four_task} compares the final performance of various EPS methods under different LLMs across four AHD problems. 
From this comparison, we draw two main observations: 
\begin{itemize}
    \setlength\itemsep{2pt}
    \item[$\circ$] There are significant variances in performance attributable to the choice of LLM for all EPS methods, with the notable exception of the OBP (OR) problem where this variance is marginal. 

    \item[$\circ$] Specifically, the EoH method shows stable and robust performance on the TSP problem across all LLMs, whereas the \baseline{}'s performance varies considerably due to its greedy nature.
\end{itemize}
\noindent 
These findings underscore the dependence of EPS methods' performance on the specific LLMs employed.


\subsubsection{Summary and Implications\\}
\mbox{}

\vspace{-5pt}
\noindent The empirical observations from previous sections jointly suggest that the LLM-based EPS algorithmic development is still in the early stages.
We hypothesize that \emph{more diverse benchmarks and applications are needed to establish a better understanding of this emergent paradigm for AHD}. 
%
Nevertheless, these preliminary results also prompt us to 
(\emph{i}) rethink the general efficacy of various components (such as prompt engineering and search strategy) within the overall paradigm, 
(\emph{ii}) consider incorporating domain knowledge to LLM-based EPS algorithm design, 
and (\emph{iii}) use a variety of LLMs to gain a more robust evaluation of the performance of EPS methods. 




\subsection{Search Cost}

\noindent \textbf{Computational Time.} We present the computation time of each independent run in Table \ref{tab:cost}. The computation time is estimated under the following settings:

\begin{enumerate}
    \item Our evaluator for the admissible set and online bin packing problems uses single-threaded evaluation, while for the TSP problem, we use eight CPU processes to perform parallel acceleration at the TSP instance level.
    
    \item We deploy and infer the open-source LLMs locally on a Tesla V100 GPU. We load the LLM's weights using \texttt{float16} precision and generate responses using the \texttt{transformers} library. We set the sampling temperature to the default value of $1.0$ and disable batch inference during text generation.
    
    \item For each independent run, we use a single evaluator and LLM. The LLM inference and function evaluation process is synchronous.
\end{enumerate}

\begin{table}[h]
    \centering
    \caption{Approximate computational cost for various AHD tasks and LLM models. We use ``CL'' and ``DS'' to denote CodeLlama and DeepSeek-Coder models respectively. 
    }
    \begin{tabular}{l|c|c|c|c|c|c|c}
    \toprule
    \multirow{2}{*}{\textbf{AHD Task}}& \multicolumn{4}{c|}{\textbf{Open source LLMs}} & \multicolumn{3}{c}{\textbf{Close source LLMs}} \\
    \cmidrule{2-8}
     & CL-7B & DS-6.7B & DS-33B & CL-34B & GPT-3.5 & GPT-4 & Claude3-Opus \\
    \midrule
    AS & \multicolumn{2}{c|}{2.5 days} & \multicolumn{2}{c|}{7 days} &  \multicolumn{3}{c}{2 days} \\
    OBP (OR) & \multicolumn{2}{c|}{2.5 days} & \multicolumn{2}{c|}{7 days} &  \multicolumn{3}{c}{2 days} \\
    OBP (Weibull) & \multicolumn{2}{c|}{3 days} & \multicolumn{2}{c|}{8 days} &  \multicolumn{3}{c}{2.5 days} \\
    TSP & \multicolumn{2}{c|}{5 days} & \multicolumn{2}{c|}{10 days} &  \multicolumn{3}{c}{4.5 days} \\
    \bottomrule
    \end{tabular}
    \label{tab:cost}
\end{table}

\vspace{-10pt}

\noindent \textbf{API Usage Pricing.}
Table \ref{tab:api_cost} shows the official API prices for different closed-source LLMs and the approximate cost of calling official APIs in each independent run. Due to price fluctuations, the API usage prices should be based on real-time prices or prices from third-party API resellers.  

\vspace{-10pt}

\begin{table}[h]
    \centering
    \caption{Approximate API prices using different closed-source LLMs.}
    \begin{tabular}{l|c|c|c}
    \toprule
     & GPT-3.5 & GPT-4 & Claude 3 Opus \\
    \midrule
    Model Input Prices (USD/K tokens)& 0.001 & 0.01 & 0.015\\
    Model Output Prices (USD/K tokens)& 0.002 & 0.03 & 0.075 \\
    Prices for Each Independent Run (USD)& 10 &  100 & 200 \\
    \bottomrule
    \end{tabular}
    
    \label{tab:api_cost}
\end{table}

\vspace{-10pt}

\noindent \textbf{Summary.} Existing LLM-based EPS methods still incur significant computational and API usage costs for the AHD task, indicating the need to conduct further optimizations and studies to accelerate existing methods and reduce the number of LLM queries.

\vspace{-10pt}

%% file: 5-conclusion.tex
\section{Conclusion}
This work presents a large-scale benchmark study comprising all existing LLM-based EPS methods along with a new proposed baseline and four AHD problems over (up-to) nine different LLMs and five independent runs. 
Based on the analyses from multiple comparison angles, we reveal novel insights into the necessity and the current progress of the LLM-based EPS paradigm for AHD.
On top of them, we summarize a few tangible implications for future research directions for LLM-based EPS, along with the fully released source codes for fostering future development.  




%% file: 6-appendix.tex
\newpage
\appendix
\section{Background Continued \label{sec:appx_A}}
\subsection{Connection to Genetic Programming}
%
%
Evolutionary program search (EPS) is conceptually similar to GP from the perspective of problem modeling (i.e., representing candidate solutions as executable computer programs). 
One of the key differences between EPS and GP lies in the representation of programs. 
GP uses relatively more abstract representations (such as tree, graph, list, etc.) to encode programs, 
while EPS directly uses executable source codes to represent programs.
Another key difference between EPS and GP lies in the creation of new solutions. 
As the name suggested, GP uses genetic operators (e.g., crossover, mutation, etc.) to generate new offspring in an explicit manner; 
while EPS utilize large language models (LLMs) to drive the search implicitly. 
This new way of using LLMs to create programs mitigates several limitations regarding GP-based approaches~\cite{o2010open}:

\begin{itemize}
\item[$\circ$] GP cannot leverage knowledge from descriptions or doc-string in natural language that describes what the program is intended to do, and neither can it generate language-based summarizations or hints to guide the following evolution process \cite{lehman2022evolution}.
However, LLMs have been trained on a number of natural language data and can easily understand the given instructions as well as do summarizations. 

\item[$\circ$] GP requires defining several problem-specific parameters, such as the function and the primitive sets. 
Designing such a set of operations is non-trivial and requires domain knowledge about the problem. 
In contrast, vast amounts of coding knowledge have been encoded within the LLM through pre-training and finetuning on an extensive unlabeled code corpus.
Therefore, LLM possesses the capability to design code akin to human-like proficiency.

\item[$\circ$] Designing effective crossover and mutation operators for increasingly complicated GP programs can be hard in practice \cite{o2010open, lehman2022evolution}. 
Whereas state-of-the-art LLMs can analyze code examples through in-context learning to generate potentially improved code.
Therefore, by combining parent programs as well as contexts with reasonable prompting strategies, LLMs are able to generate more diverse and effective programs.
\end{itemize}

\noindent Therefore, while modeling problems as computer programs are initially pioneered by GP, the advent of LLM has significantly enhanced the ability to search under this representation paradigm.

\subsection{Existing EPS Methods}
\textbf{FunSearch}~\cite{romera2024mathematical} evolves heuristics for mathematical and combinatorial optimization problems, achieving superior results compared to existing solutions on the cap set~\cite{grochow2019new} and admissible set problems~\cite{tao2006additive}. 
The input to FunSearch consists of a code template (called ``specification''), which defines a template heuristic to be evolved and a function for evaluating the searched heuristics. 
FunSearch employs an island-based evolution strategy with restart. 
Specifically, the island-based population comprises multiple islands. 
Each island incorporates several clusters, each of which collects heuristics with the same fitness scores. 

During the evolutionary process, FunSearch selects a random island and chooses $k$ clusters (defaulting by two), then selects one function within each cluster. 
In this process, shorter heuristic functions (with fewer lines) within the cluster are preferred. FunSearch then samples $N$ new heuristic functions based on these heuristic examples. 
The predefined evaluator then evaluates the newly obtained heuristics, and the valid (no errors or timeout in evaluation) heuristic functions are then registered back to the same island. 
Periodically, FunSearch removes half of the individuals from the worst-performing island (where the best function has the worst fitness score) 
and replaces them with individuals from other islands that perform better.

FunSearch utilizes few-shot prompt engineering while sampling new heuristics from LLM. 
Specifically, the selected heuristic functions are renamed as \emph{function\_name\_v0}, \emph{function\_name\_v1}, ..., \emph{function\_name\_vk} in order of increasing fitness scores, providing an empty function call (function declaration without a function body) \emph{def function\_name\_v(k+1)} to let LLM complete the code. 
Therefore, code completion models, such as StarCoder~\cite{li2023starcoder}, can be used by FunSearch.

\vspace{3pt}
\noindent\textbf{Evolution of Heuristic (EoH)}~\cite{liu2023algorithm}  evolves heuristics for solving combinatorial optimization problems, consequently outperforms heuristics generated by automatic heuristic design (AHD) on traveling salesman (TSP)~\cite{matai2010traveling} and online bin packing problems~\cite{seiden2002online}. 
The input to the EoH is a task description including a brief description of the AHD task, the input/output of the heuristic function, and the goal of the heuristic function.

EoH employs the GA strategy, the initial population is obtained by sampling from the LLM based on the task description, without the necessity of specifying a template heuristic function.
EoH proposes multiple crossover and mutation strategies through various prompt settings, such as exploring new heuristics that are completely different from existing heuristics, exploring new heuristics that are based on the observation of existing heuristics, modifying the parameters for one heuristic, etc.

EoH uses Chain of Thought (CoT) \cite{wei2022chain} prompt engineering and evolves not only code but also thoughts of the heuristic function. 
The prompt content of EoH is composed of \ding{172} genetic operator-related instructions, \ding{173} selected heuristic functions (serve as ``parents'' in crossover), and \ding{174} corresponded ``thoughts'' of each selected heuristic function. 
During generation, LLM is asked not only to implement the code of the function but also to describe the thought and the idea of the code. 

\vspace{3pt}
\noindent\textbf{Reflective Evolution (ReEvo)}~\cite{ye2024reevo} evolves heuristics for solving combinatorial optimization problems. 
ReEvo requires a template heuristic function and a task description as the input. 

Similar to EoH, ReEvo uses a GA framework. During initialization, the population is filled by performing mutation and crossover operators to the template heuristic. 
The reflection mechanisms \cite{brown2020language} are utilized in its crossover and mutation operators. 
ReEvo proposes two types of reflection mechanisms, i.e., short-term reflection and long-term reflection to let LLM analyze historical context.
The short-term reflection initially allows LLM to analyze the design rationale of heuristic parents and provide generated hints, which are then encapsulated into prompts to enable LLM to generate new individuals.
The long-term reflection summarizes clues generated by multiple short-term reflections and guides the mutation operator.

\vspace{3pt}
\noindent\textbf{Eureka}~\cite{ma2024eureka} evolves reward function used in reinforcement learning, consequently outperforms human experts on most tasks in tested RL environments. 
Eureka is completely free of task-specific prompts, reward templates, as well as few-shot examples. 
Eureka takes the source code of the RL environment and task description as context, and zero-shot generates executable reward functions from the LLM. 
Similar to EoH, multiple crossover and mutation operators are proposed for editing reward functions, such as changing the hyperparameter of existing reward components, changing the functional form of existing reward components, and introducing new reward components.
And a reward reflection is also performed after evaluation to obtain feedback on how to optimize the reward function design.

\section{Template Heuristics\label{sec:appx_C}}
The template heuristic for the admissible set problem is illustrated in Listing~\ref{lst:admi_seed}. Following the prior work \cite{romera2024mathematical}, we employ a ``dummy'' template heuristic that simply returns a constant value. This inherently minimizes the influence of expert knowledge in the design of the template heuristic function.

\begin{lstlisting}[caption={Template heuristic for admissible set problem.}, frame=tlrb,language=Python, label={lst:admi_seed}]{Name}
def priority(el: tuple[int, ...], n: int, w: int) -> float:
    """Returns the priority with which we want to add `el' 
    to the set.
    Args:            
        el: A vector represents possible element of the set.
        n : The dimension of the `el' vector.
        w : Number of non-zero elements in the `el' vector.
    Return:
        The priority of the `el' vector.
    """
    return 0.0
\end{lstlisting}

The template heuristic for the online bin packing problem and the traveling salesman problem is shown in Listing \ref{lst:bin_seed} and Listing \ref{lst:tsp_seed}. 
Since both functions include array-type arguments and return values, it is challenging to design a simple and valid template for relatively complex template heuristic functions.
Therefore, these templates are generated randomly by a large language model (LLM) based on its function declaration and docstring.
In our benchmark, we make all LLM-based EPS methods use the same template for the same AHD problem to guarantee fairness.

\newpage
\begin{lstlisting}[caption={Template heuristic for online bin packing (OR and Weibull).}, frame=tlrb,language=Python, label={lst:bin_seed}]{Name}
import numpy as np

def priority(item: float, bins: np.ndarray) -> np.ndarray:
    """Returns priority with which we want to add the item 
    to each bin.
    Args:
        item: Size of item to be added to the bin.
        bins: Array of capacities for each bin.
    Return:
        An array of the same size as bins with priority score 
        of each bin.
    """
    penalty = np.arange(len(bins), 0, -1)
    scores = bins / (bins - item) - penalty
    max_capacity_bins = np.where(bins == bins.max())[0]
    for idx in max_capacity_bins:
        scores[idx] = -np.inf
    return scores
 \end{lstlisting}

\begin{lstlisting}[caption={Template heuristic for travelling salesman problem.}, frame=tlrb,language=Python, label={lst:tsp_seed}]{Name}
import numpy as np

def priority(score_mat: np.ndarray, loc_opt: np.ndarray, 
             edge_used:np.ndarray) -> np.ndarray:
    """Update the score of each edge in score_mat.
    Args:
        score_mat: Score matrix, each element represents 
                   the `score' of each edge.
        loc_opt  : Local optimal solution path.
        edge_used: Matrix representing the number of 
                   times each edge is used.
    Return:
        updated_score_mat: the updated score_mat
    """
    num_nodes = score_mat.shape[0]
    updated_score_mat = np.copy(score_mat)
    for i in range(num_nodes - 1):
        cur_node = loc_opt[i]
        next_node = loc_opt[i + 1]
        updated_score_mat[cur_node, next_node] *= \
            (1 + edge_used[cur_node, next_node])
        updated_score_mat[loc_opt[-1], loc_opt[0]] *= \
            (1 + edge_used[loc_opt[-1], loc_opt[0]])
    return updated_score_mat
 \end{lstlisting}

\section{More Results and Analysis\label{sec:appx_B}}
\subsection{Ablation Study for EoH under Different Experimental Settings \label{sec:appx_expt1}}
In this section, we present an ablation study on the different experimental settings of EoH. 
As mentioned in Sec. \S\ref{sec:prelim}, we modified the default EoH settings to standardize the initialization method across all EPS methods:
\begin{itemize}
    \item The template program is not required for the default EoH settings, as it fills the population by sampling from LLM according to the task-specific prompt.
    In our benchmark, we initially incorporate a template heuristic into the initial population, and repeatedly apply crossover and mutation operators to the existing individuals in the population to generate new individuals until the desired population size is achieved.
    
    \item We also increase the default population size from 20 to 100 considering that more sampled heuristics are obtained (2,000 in default EoH settings and 10,000 in our benchmark).
\end{itemize} 

\noindent Table \ref{tab:ablation} compares the performance using different experimental settings. We use EoH-D to denote EoH with default settings, where the template heuristic is not provided during initialization, and the population size is set to 20. However, we maintain the maximum number of sampled heuristics to 10,000 to promise consistency with our benchmark settings.
We notice that the performance variation is notable in the admissible set problem, while it is marginal in the other problems. 
This suggests that applying the default settings, i.e., initializing the population by fully sampling from LLM, and a relatively smaller population size, should be a more effective setting for EoH.

\begin{table}[h]
    \centering
    \caption{The performance of the top-1 heuristics achieved by two EoH settings with 10,000 query budgets. We use EoH and EoH-D to denote the benchmark settings in this work,  and the default settings in EoH's paper, respectively. The performance is measured as the mean relative distance to the best-known optimum ($\Delta_{\rm d}$) averaged over five independent runs using the GPT-3.5 model. Lower $\Delta_{\rm d}$ indicates better performance.}
    \begin{tabular}{l|c|c|c}
    \toprule
    EoH Setting & Admissible Set & Online Bin Packing & Travelling Salesman ($\times10^{-2}$) \\
    \midrule
    EoH   &7.89 & 3.81 & 3.26 \\
    EoH-D    &6.85 & 3.82 & 3.20 \\ 
    \bottomrule
    \end{tabular}
    \label{tab:ablation}
\end{table}

\begin{figure}[ht!]
     \centering
     \begin{subfigure}[b]{0.98\textwidth}
         \centering
         \includegraphics[width=\textwidth]{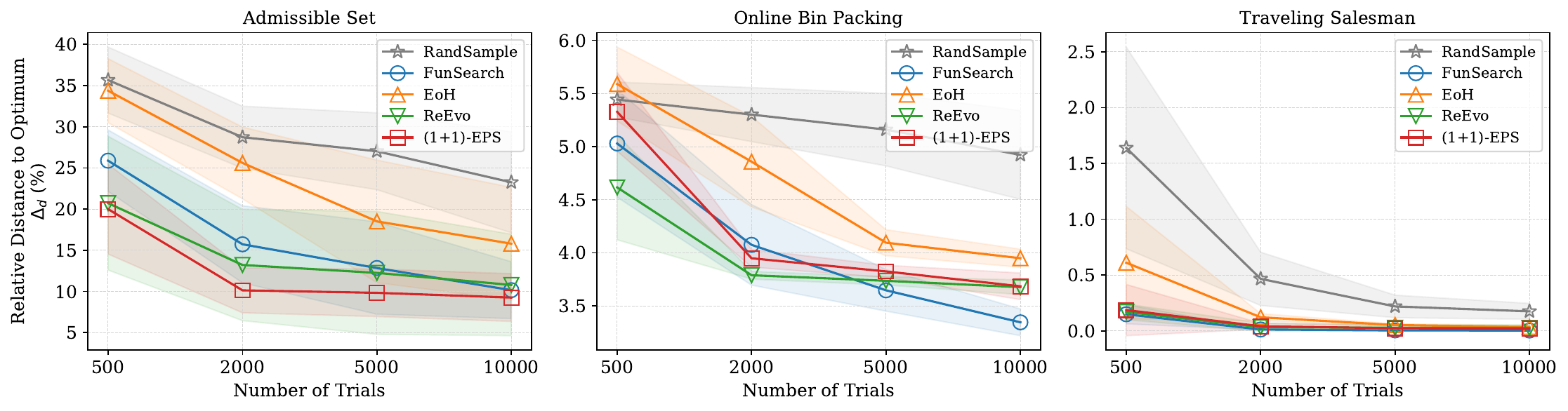}
         \caption{CodeLlama-7B}
     \end{subfigure}\\
     \begin{subfigure}[b]{0.98\textwidth}
         \centering
         \includegraphics[width=\textwidth]{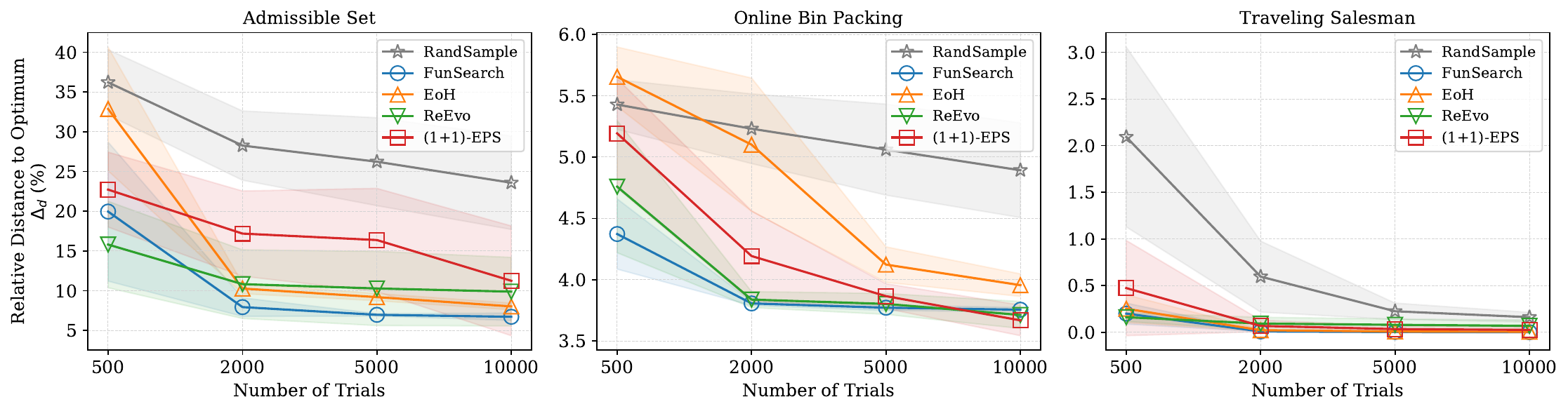}
         \caption{CodeLlama-34B}
     \end{subfigure}\\
          \begin{subfigure}[b]{0.98\textwidth}
         \centering
         \includegraphics[width=\textwidth]{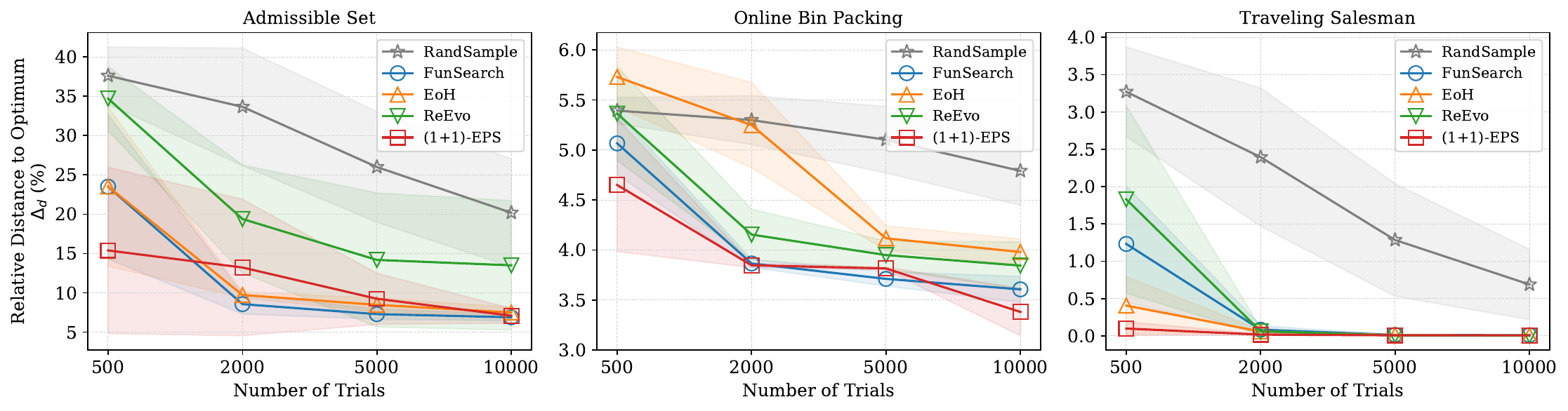}
         \caption{DeepSeek-Coder-6.7B}
     \end{subfigure}\\
          \begin{subfigure}[b]{0.98\textwidth}
         \centering
         \includegraphics[width=\textwidth]{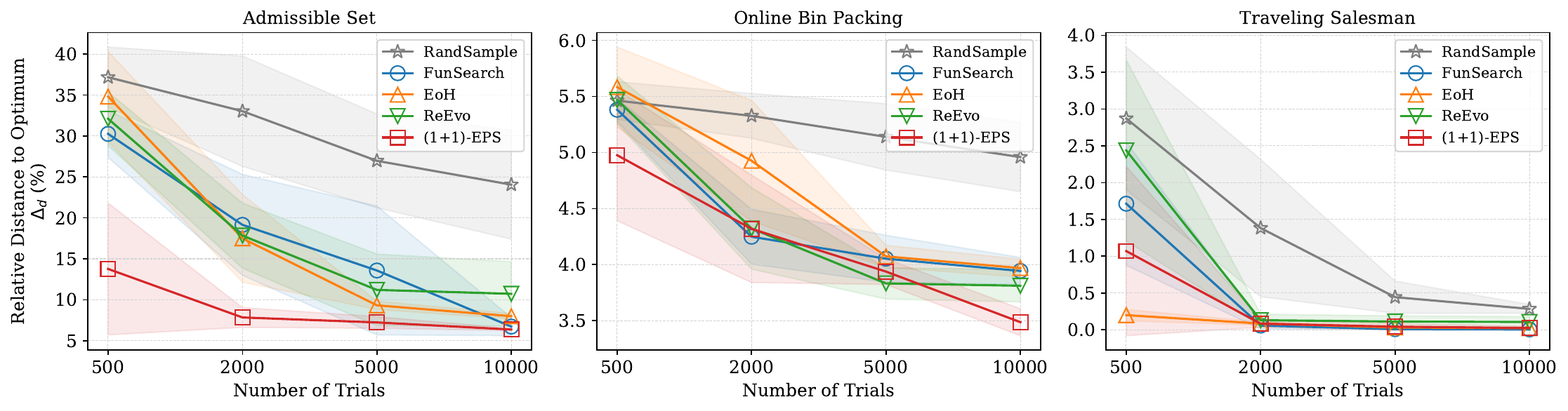}
         \caption{DeepSeek-Coder-33B}
     \end{subfigure}
     
    \caption{Convergence curve comparison on the performance of the top-$5$\textperthousand{} heuristics generated by various LLMs. The mean relative distance to the best-known optimum ($\Delta_{\rm d}$) aggregated over five independent runs are denoted with markers, while the standard deviations of $\Delta_{\rm d}$ are highlighted with the shaded regions.}
    \label{fig:appx_convergence}
\end{figure}

\vspace{-2em}

\begin{figure}[h!]
     \centering
     \begin{subfigure}[b]{0.98\textwidth}
         \centering
         \includegraphics[width=\textwidth]{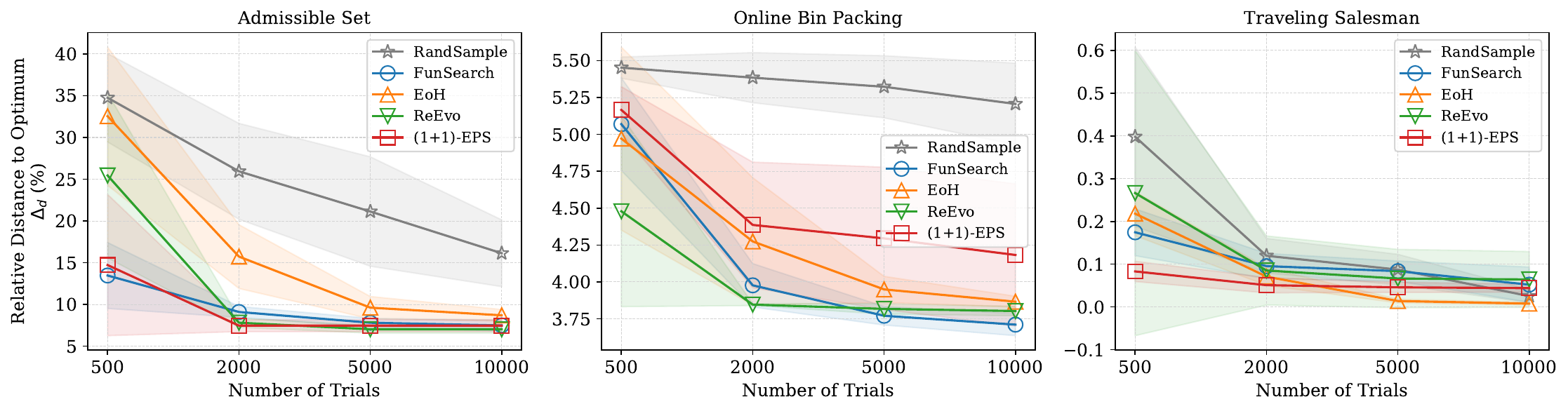}
         \caption{GPT-3.5}
     \end{subfigure}\\
     \begin{subfigure}[b]{0.98\textwidth}
         \centering
         \includegraphics[width=\textwidth]{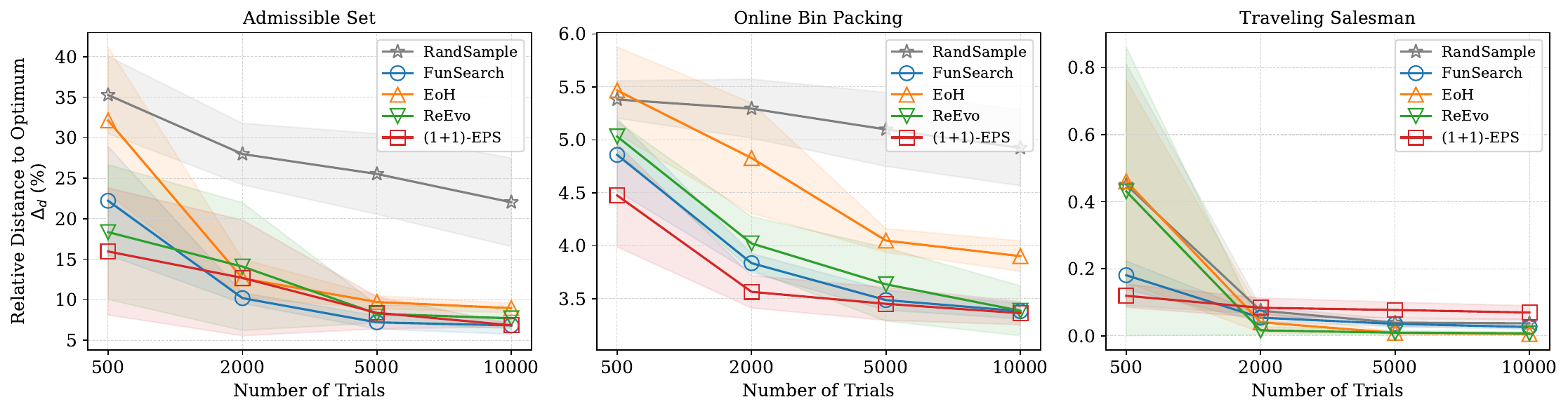}
         \caption{GPT-4}
     \end{subfigure}\\
          \begin{subfigure}[b]{0.98\textwidth}
         \centering
         \includegraphics[width=\textwidth]{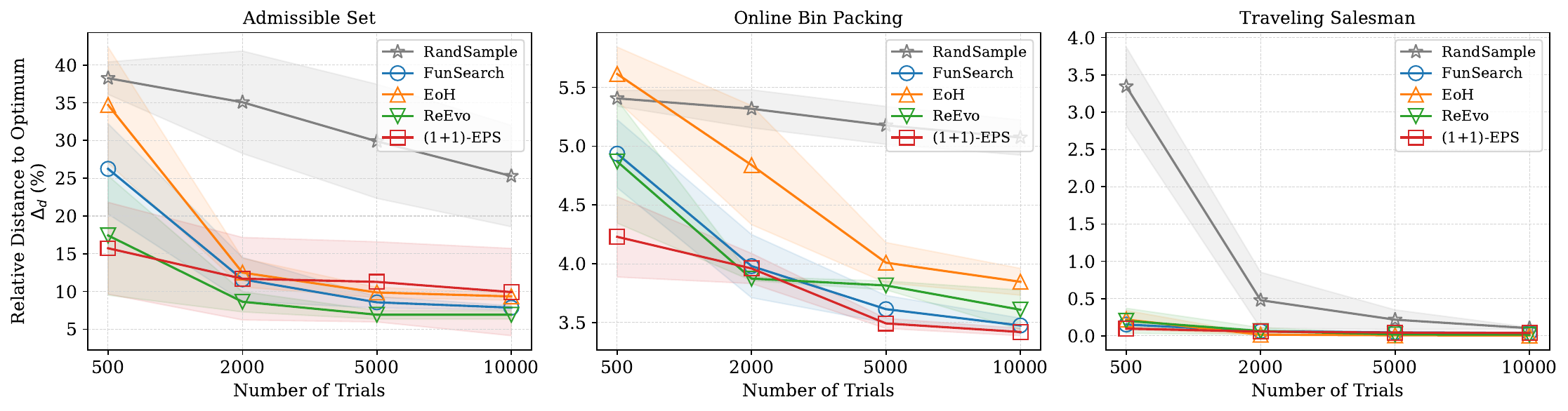}
         \caption{Claude 3 Opus}
     \end{subfigure}
     
    \caption{Convergence curve comparison on the performance of the top-$5$\textperthousand{} heuristics generated by various LLMs. The mean relative distance to the best-known optimum ($\Delta_{\rm d}$) aggregated over five independent runs are denoted with markers, while the standard deviations of $\Delta_{\rm d}$ are highlighted with the shaded regions.}
    \label{fig:appx_convergence1}
\end{figure}

\begin{figure}[t]
     \centering
     \begin{subfigure}[b]{0.95\textwidth}
         \centering
         \includegraphics[width=\textwidth]{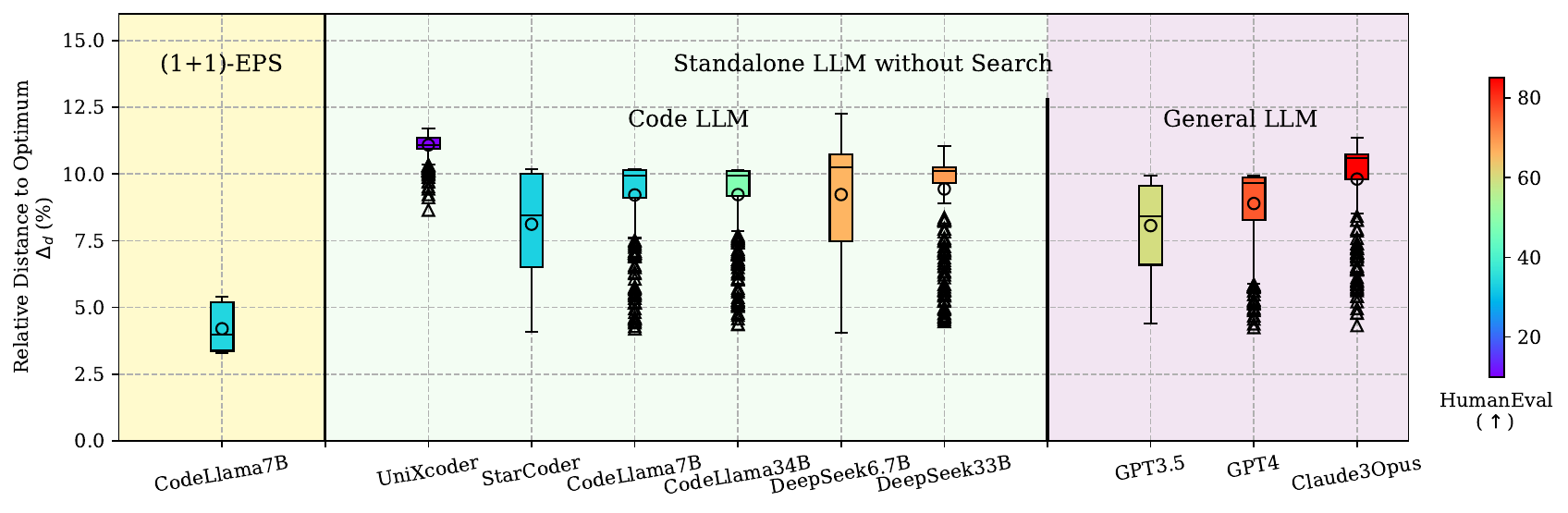}
         \label{fig:llm_alone_top_1_100_model_choice}
     \end{subfigure}
    \caption{Box plot comparison on the performance of the top-$1\%$ heuristics generated by LLMs with varying capacities under 10,000 query budgets. 
    We group LLMs into two categories: (1) LLMs specialized for coding tasks (with background shaded in \prioritized[lighgreenbackground]) and (2) general-purpose LLMs (with background shaded in \prioritized[lighpurplebackground]).
    Then, the LLMs are arranged in the order of ascending model size within each group. 
    The color scale of the boxes corresponds with the scores on HumanEval \cite{chen2021evaluating}.
    The performance is measured as the relative distance to the best-known optimum ($\Delta_{\rm d}$) aggregated over four AHD problems and five independent runs. Lower $\Delta_{\rm d}$ indicates better performance. 
    The performance of the simple baseline \baseline{} with CodeLlama-7B is also provided as a reference.
    }
    \label{fig:llm_alone_model_choice_top_1_100}
\end{figure}

\begin{figure}[ht!]
     \centering
     \begin{subfigure}[b]{0.98\textwidth}
         \centering
         \includegraphics[width=\textwidth]{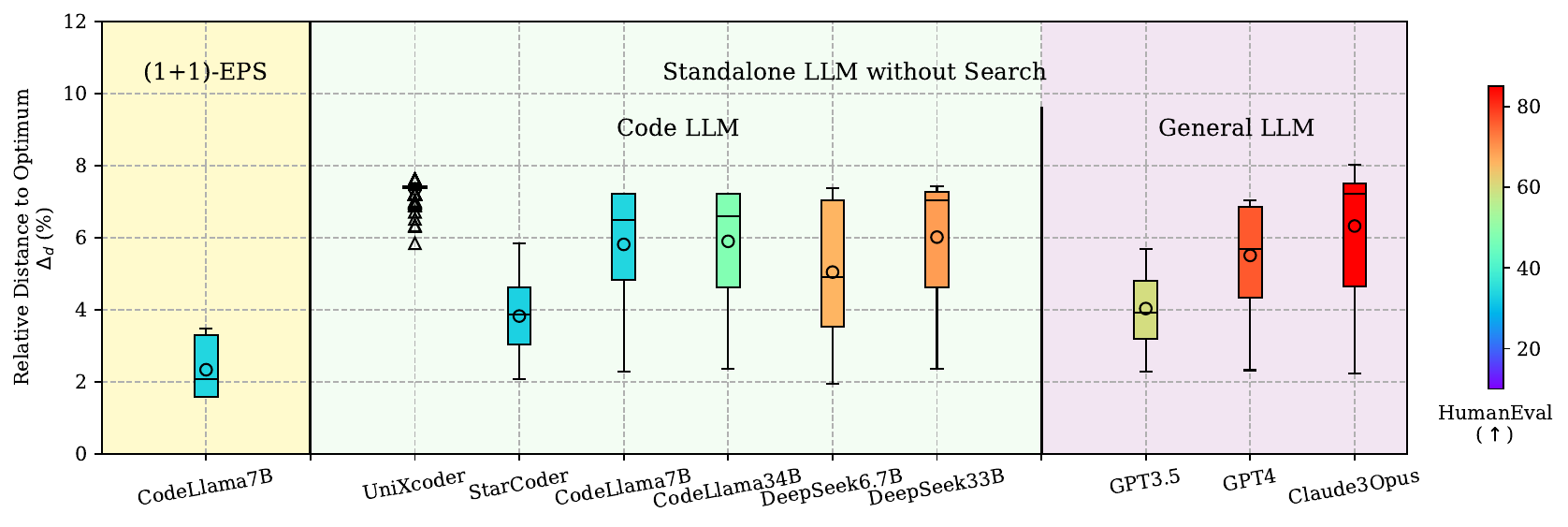}
         \caption{Admissible Set}
     \end{subfigure}\\
     \begin{subfigure}[b]{0.98\textwidth}
         \centering
         \includegraphics[width=\textwidth]{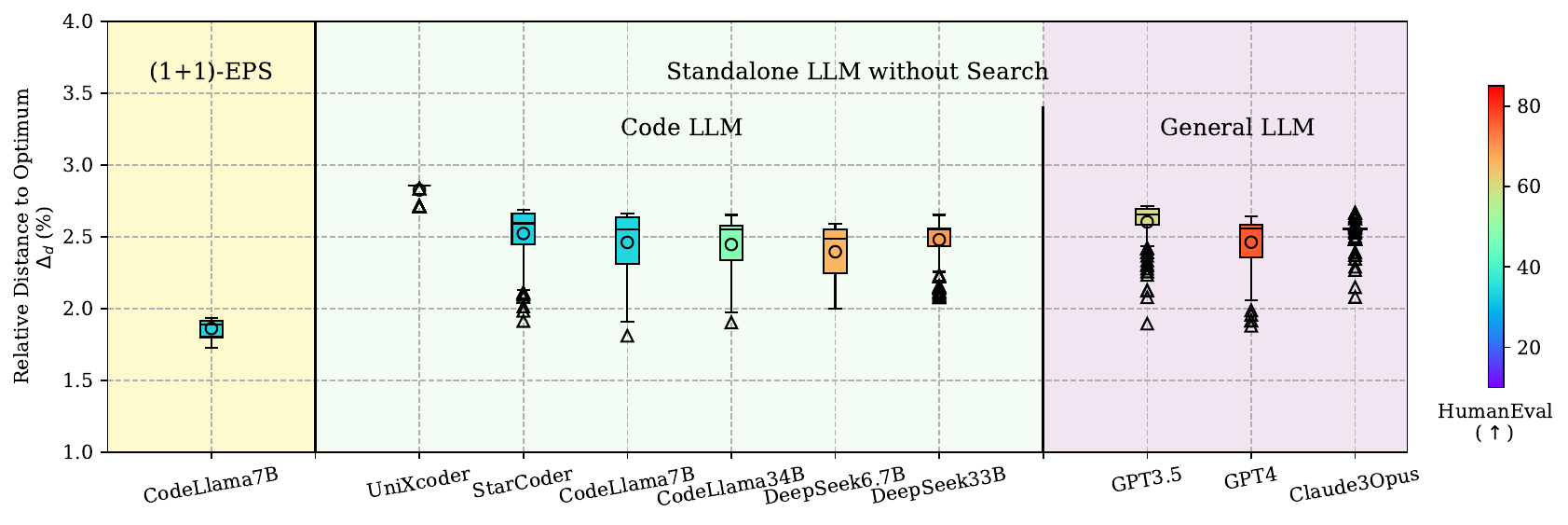}
         \caption{Online Bin Packing}
     \end{subfigure}\\
          \begin{subfigure}[b]{0.98\textwidth}
         \centering
         \includegraphics[width=\textwidth]{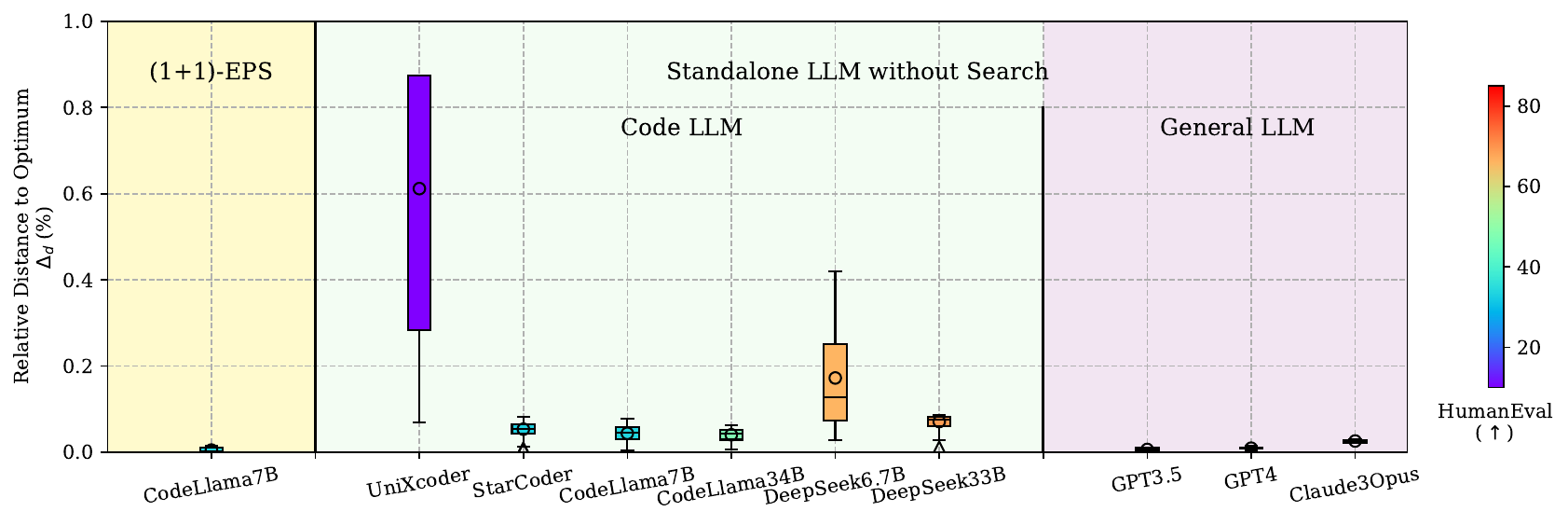}
         \caption{Traveling Salesman}
     \end{subfigure}
     
    \caption{Box plot comparison on the performance of the top-$5$\textperthousand{} heuristics generated by LLMs with varying capacities under 10,000 query budgets. The performance is measured as the relative distance to the best-known optimum ($\Delta_{\rm d}$) aggregated over four AHD problems and five independent runs. Lower $\Delta_{\rm d}$ indicates better performance. We demonstrate detailed comparison results on (a) admissible set, (b) online bin packing, and (c) traveling salesman problems.}
    \label{fig:llm_choices_top_5_1000_individual}
\end{figure}

\subsection{More Results on the Impact of Query Budget \label{sec:appx_expt2}}

This section is a continuation of Sec. \S\ref{subsubsec:expt1_angle1}, where we demonstrate detailed results on individual AHD problems and LLM choices under different query budges.
In Fig.~\ref{fig:appx_convergence} and Fig.~\ref{fig:appx_convergence1}, we visualize the performance of the top-$5$\textperthousand{} heuristics achieved by standalone LLM under different query budgets. The standalone LLM is denoted as ``RandomSample'' in the caption. We also provide the results of various LLM-based EPS for comparison. 

Extensive results on different LLMs and AHD problems demonstrate that while the performance of standalone LLMs on AHD problems generally improves with increasing query budgets, several observations are also obtained:

\begin{itemize}
    \item We observe a noticeable performance gap between standalone LLM and EPS methods across various AHD problems and LLM choices. 
    By using only 2,000 query budgets, the performance of the heuristics obtained by EPS methods can approach or even exceed that of the standalone LLMs using 10,000 query budgets.
    This further underscores the significance of integrating evolutionary search methods with LLMs in AHD tasks. 
    
    \item The heuristic obtained by (1+1)-EPS with 500 query budgets can outperform standalone LLM with 10,000 query budgets under most LLM choices and AHD tasks.
    
    \item We find that after using 2,000 query budgets, the convergence rate of the EPS methods slows down, which is particularly evident in the traveling salesman problem. 
\end{itemize}

\subsection{Performance in Each AHD Problem and LLM Choice \label{sec:appx_expt3}}
This section is a continuation of Sec. \S\ref{subsubsec:expt1_angle2}, we present more elaborated results comparing the performance of various LLM choices on individual AHD problems. 
Fig. \ref{fig:llm_alone_model_choice_top_1_100} demonstrates the aggregated performance of the top-1\% heuristics generated by LLMs with varying capacities. 
The performance proposed baseline (1+1)-EPS with CodeLlama-7B is provided as a reference. In Fig. \ref{fig:llm_choices_top_5_1000_individual}, we provide the performance of top-5\textperthousand{} heuristics generated by different LLMs on individual AHD problems. 

Fig. \ref{fig:llm_alone_model_choice_top_1_100} and Fig. \ref{fig:llm_choices_top_5_1000_individual} further support the results presented in Sec. \S\ref{subsubsec:expt1_angle2} that LLMs with more capacity do not promise better performance on AHD problems, coupling a relatively small capacity LLM (i.e., CodeLlama-7B) with search strategy can make a difference. 
We draw two new observations from the comparison results on three AHD problems in Fig. \ref{fig:llm_choices_top_5_1000_individual} that:
\begin{itemize}
    \item In the online bin packing problem, the performance differences between different LLMs are minimal. However, in the admissible set and especially in traveling salesman problems, the disparities are more explicit.
    \item No single LLM demonstrates significantly superior performance across all tasks. But conventional LM, such as UniXcoder, is consistently inferior to modern LLMs on these three AHD tasks.
\end{itemize}

\subsection{Top-1 Performance Results}
In this section, we present extensive experimental results comparing the performance of the top-1 heuristics obtained by standalone LLMs and four LLM-based EPS methods on four AHD tasks and seven LLM models.
The performance is measured by the relative distance to the best-known optimum ($\Delta_{\rm d}$, lower is better). 
Table \ref{tab:top1_mean_std} lists the average and standard deviation on the performance of top-1 heuristics over five independent runs.
Table \ref{tab:top1_best} showcases the best performance on the top-1 heuristics over five independent runs.
The standalone LLM is denoted by ``RandSample'' in these tables. 
We can conclude from both Table \ref{tab:top1_mean_std} and Table \ref{tab:top1_best} that: 
\begin{itemize}
    \item Based on the mean and standard deviation results among five independent runs (shown in Table \ref{tab:top1_mean_std}), the LLM-based EPS methods generally outperform the standalone LLM in the AS, OBP (Weibull), and TSP problems, with the disparity being particularly notable in the TSP problem. However, in the OBP (OR) problem, the LLM-based EPS methods do not demonstrate a significant advantage.

    \item Similarly, the results based on the best performance among five independent runs (shown in Table \ref{tab:top1_best}) also indicate that the LLM-based EPS methods significantly outperform the standalone LLM in the AS, OBP (Weibull), and TSP problems, but do not show a clear advantage in the OBP (OR) problem.
    
    \item With a maximum query budget of 10,000, none of the existing EPS methods demonstrate significant advantages over standalone LLMs in specific AHD tasks, indicating the necessity of developing more effective EPS methods. 
\end{itemize}

\begin{table}[t]
    \centering
    \caption{The performance of the top-1 heuristics. The mean and standard deviation of the relative distance to the best-known optimum ($\Delta_{\rm d}$) over five independent runs are reported. Lower $\Delta_{\rm d}$ indicates better performance. 
    We use ``CL'' and ``DS'' to denote the CodeLlama and DeepSeek model respectively.
    We use ``AS'', ``OR'', ``WEI'', and ``TSP'' to denote admissible set, online bin packing (OR), online bin packing (Weibull), and traveling salesman problem, respectively.
    }
\resizebox{0.98\textwidth}{!}{
    \begin{tabular}{c|c|c|c|c|c|c|c|c}
    \toprule
Task & Method & CL-7B & CL-34B & DS-6.7B & DS-33B & GPT-3.5 & GPT-4 & Claude 3 \\ \midrule
\multirow{5}{*}{\makecell{AS}} & RandSample & ${ 9.66}_{\pm 0.73}$ & ${ 10.06}_{\pm 0.80}$ & ${ 8.82}_{\pm 0.76}$ & ${ 9.96}_{\pm 0.37}$ & ${ 9.89}_{\pm 0.62}$ & ${ 10.12}_{\pm 0.79}$ & ${ 9.79}_{\pm 0.67}$ \\
& FunSearch & ${ 10.16}_{\pm 3.45}$ & ${ 6.73}_{\pm 0.46}$ & ${ 6.89}_{\pm 0.43}$ & ${ 6.49}_{\pm 1.02}$ & ${ 7.33}_{\pm 0.52}$ & ${ 6.59}_{\pm 0.78}$ & ${ 7.49}_{\pm 0.36}$ \\
& EoH & ${ 14.09}_{\pm 5.43}$ & ${ 7.26}_{\pm 0.90}$ & ${ 6.83}_{\pm 0.33}$ & ${ 7.59}_{\pm 0.28}$ & ${ 7.89}_{\pm 0.08}$ & ${ 7.83}_{\pm 0.69}$ & ${ 8.33}_{\pm 0.45}$ \\
& ReEvo & ${ 10.79}_{\pm 6.16}$ & ${ 9.89}_{\pm 4.31}$ & ${ 11.46}_{\pm 6.12}$ & ${ 10.72}_{\pm 3.93}$ & ${ 7.03}_{\pm 0.46}$ & ${ 7.16}_{\pm 0.63}$ & ${ 6.93}_{\pm 0.58}$ \\
& (1+1)-EPS & ${ 9.26}_{\pm 2.90}$ & ${ 11.26}_{\pm 6.91}$ & ${ 7.09}_{\pm 0.94}$ & ${ 6.36}_{\pm 0.12}$ & ${ 7.46}_{\pm 0.71}$ & ${ 6.86}_{\pm 0.21}$ & ${ 9.39}_{\pm 5.02}$ \\

 \midrule
\multirow{5}{*}{\makecell{OR}} & RandSample & ${ 6.64}_{\pm 0.21}$ & ${ 6.78}_{\pm 0.08}$ & ${ 6.84}_{\pm 0.13}$ & ${ 6.98}_{\pm 0.00}$ & ${ 6.71}_{\pm 0.13}$ & ${ 6.67}_{\pm 0.00}$ & ${ 6.91}_{\pm 0.05}$ \\
& FunSearch & ${ 5.89}_{\pm 0.27}$ & ${ 6.84}_{\pm 0.05}$ & ${ 6.54}_{\pm 0.21}$ & ${ 6.98}_{\pm 0.00}$ & ${ 6.37}_{\pm 0.08}$ & ${ 6.09}_{\pm 0.19}$ & ${ 6.19}_{\pm 0.13}$ \\
& EoH & ${ 6.91}_{\pm 0.10}$ & ${ 6.88}_{\pm 0.08}$ & ${ 6.88}_{\pm 0.15}$ & ${ 6.95}_{\pm 0.05}$ & ${ 6.95}_{\pm 0.05}$ & ${ 6.61}_{\pm 0.26}$ & ${ 6.78}_{\pm 0.15}$ \\
& ReEvo & ${ 6.61}_{\pm 0.19}$ & ${ 6.64}_{\pm 0.13}$ & ${ 6.74}_{\pm 0.34}$ & ${ 6.74}_{\pm 0.34}$ & ${ 6.95}_{\pm 0.05}$ & ${ 5.99}_{\pm 0.43}$ & ${ 6.50}_{\pm 0.40}$ \\
& (1+1)-EPS & ${ 6.47}_{\pm 0.29}$ & ${ 6.47}_{\pm 0.29}$ & ${ 6.02}_{\pm 0.51}$ & ${ 6.02}_{\pm 0.51}$ & ${ 6.78}_{\pm 0.15}$ & ${ 5.92}_{\pm 0.13}$ & ${ 6.06}_{\pm 0.08}$ \\

 \midrule
\multirow{5}{*}{\makecell{WEI}} & RandSample & ${ 0.99}_{\pm 0.08}$ & ${ 1.07}_{\pm 0.21}$ & ${ 1.33}_{\pm 0.01}$ & ${ 1.34}_{\pm 0.01}$ & ${ 1.43}_{\pm 0.37}$ & ${ 0.94}_{\pm 0.07}$ & ${ 1.75}_{\pm 0.32}$ \\
& FunSearch & ${ 0.64}_{\pm 0.04}$ & ${ 0.63}_{\pm 0.03}$ & ${ 0.64}_{\pm 0.01}$ & ${ 0.77}_{\pm 0.06}$ & ${ 0.72}_{\pm 0.05}$ & ${ 0.63}_{\pm 0.03}$ & ${ 0.72}_{\pm 0.09}$ \\
& EoH & ${ 0.75}_{\pm 0.01}$ & ${ 0.73}_{\pm 0.03}$ & ${ 0.74}_{\pm 0.04}$ & ${ 0.69}_{\pm 0.06}$ & ${ 0.67}_{\pm 0.03}$ & ${ 0.70}_{\pm 0.03}$ & ${ 0.69}_{\pm 0.01}$ \\
& ReEvo & ${ 0.71}_{\pm 0.05}$ & ${ 0.70}_{\pm 0.01}$ & ${ 0.92}_{\pm 0.27}$ & ${ 0.87}_{\pm 0.07}$ & ${ 0.65}_{\pm 0.03}$ & ${ 0.76}_{\pm 0.07}$ & ${ 0.64}_{\pm 0.04}$ \\
& (1+1)-EPS & ${ 0.83}_{\pm 0.23}$ & ${ 0.80}_{\pm 0.08}$ & ${ 0.67}_{\pm 0.06}$ & ${ 0.89}_{\pm 0.20}$ & ${ 1.56}_{\pm 1.03}$ & ${ 0.79}_{\pm 0.08}$ & ${ 0.68}_{\pm 0.05}$ \\

 \midrule
\multirow{5}{*}{\makecell{TSP\\($\times 10^{-2}$)}} & RandSample & ${ 1.97}_{\pm 0.00}$ & ${ 5.31}_{\pm 0.02}$ & ${ 13.69}_{\pm 0.03}$ & ${ 10.37}_{\pm 0.04}$ & ${ 2.19}_{\pm 0.02}$ & ${ 2.88}_{\pm 0.01}$ & ${ 9.64}_{\pm 0.01}$ \\
& FunSearch & ${ 0.33}_{\pm 0.00}$ & ${ 0.08}_{\pm 0.00}$ & ${ 0.37}_{\pm 0.00}$ & ${ 0.38}_{\pm 0.00}$ & ${ 4.16}_{\pm 0.04}$ & ${ 1.73}_{\pm 0.00}$ & ${ 3.54}_{\pm 0.02}$ \\
& EoH & ${ 2.28}_{\pm 0.01}$ & ${ 0.22}_{\pm 0.00}$ & ${ 0.23}_{\pm 0.00}$ & ${ 1.20}_{\pm 0.01}$ & ${ 0.33}_{\pm 0.00}$ & ${ 0.18}_{\pm 0.00}$ & ${ 0.12}_{\pm 0.00}$ \\
& ReEvo & ${ 2.00}_{\pm 0.02}$ & ${ 5.45}_{\pm 0.06}$ & ${ 0.26}_{\pm 0.00}$ & ${ 8.64}_{\pm 0.06}$ & ${ 5.80}_{\pm 0.06}$ & ${ 0.46}_{\pm 0.00}$ & ${ 0.98}_{\pm 0.00}$ \\
& (1+1)-EPS & ${ 1.80}_{\pm 0.02}$ & ${ 1.73}_{\pm 0.02}$ & ${ 0.70}_{\pm 0.00}$ & ${ 1.82}_{\pm 0.02}$ & ${ 3.97}_{\pm 0.00}$ & ${ 5.75}_{\pm 0.03}$ & ${ 3.87}_{\pm 0.01}$ \\
    \bottomrule
    \end{tabular}
}
    \label{tab:top1_mean_std}
\end{table}

\begin{table}[t]
    \centering
    \caption{The performance of the top-1 heuristics. The minimum of the relative distance to the best-known optimum ($\Delta_{\rm d}$) over five independent runs is reported. 
    Lower $\Delta_{\rm d}$ indicates better performance. 
    We use ``CL'' and ``DS'' to denote the CodeLlama and DeepSeek model respectively. 
    We use ``AS'', ``OR'', ``WEI'', and ``TSP'' to denote admissible set, online bin packing (OR), online bin packing (Weibull), and traveling salesman problem, respectively.
    }
\resizebox{.92\textwidth}{!}{
    \begin{tabular}{c|c|c|c|c|c|c|c|c}
    \toprule
Task & Method & CL-7B & CL-34B & DS-6.7B & DS-33B & GPT-3.5 & GPT-4 & Claude 3 \\ \midrule
\multirow{5}{*}{\makecell{AS}} & RandSample &  9.09 &  9.49 &  7.79 &  9.49 &  9.09 &  9.29 &  8.89 \\
& FunSearch &  6.49 &  6.09 &  6.49 &  5.09 &  6.59 &  5.59 &  6.99 \\
& EoH &  7.29 &  5.99 &  6.39 &  7.19 &  7.79 &  6.99 &  7.69 \\
& ReEvo &  5.89 &  6.69 &  6.59 &  7.79 &  6.39 &  6.29 &  6.29 \\
& (1+1)-EPS &  6.29 &  5.59 &  5.99 &  6.19 &  6.49 &  6.59 &  4.50 \\

 \midrule
\multirow{5}{*}{\makecell{OR}} & RandSample &  6.37 &  6.67 &  6.67 &  6.98 &  6.57 &  6.67 &  6.88 \\
& FunSearch &  5.65 &  6.78 &  6.26 &  6.98 &  6.26 &  5.95 &  6.06 \\
& EoH &  6.78 &  6.78 &  6.67 &  6.88 &  6.88 &  6.26 &  6.57 \\
& ReEvo &  6.47 &  6.47 &  6.26 &  6.26 &  6.88 &  5.54 &  5.95 \\
& (1+1)-EPS &  6.26 &  6.26 &  5.34 &  5.34 &  6.67 &  5.75 &  5.95 \\

 \midrule
\multirow{5}{*}{\makecell{WEI}} & RandSample &  0.88 &  0.84 &  1.32 &  1.32 &  0.91 &  0.85 &  1.44 \\
& FunSearch &  0.59 &  0.59 &  0.63 &  0.71 &  0.68 &  0.59 &  0.63 \\
& EoH &  0.73 &  0.69 &  0.70 &  0.64 &  0.63 &  0.68 &  0.68 \\
& ReEvo &  0.65 &  0.69 &  0.67 &  0.77 &  0.62 &  0.67 &  0.60 \\
& (1+1)-EPS &  0.65 &  0.73 &  0.60 &  0.68 &  0.76 &  0.73 &  0.62 \\

 \midrule
\multirow{5}{*}{\makecell{TSP\\($\times 10^{-2}$)}} & RandSample &  1.56 &  2.38 &  11.26 &  4.47 &  0.56 &  1.40 &  8.78 \\
& FunSearch &  0.14 &  0.03 &  0.19 &  0.29 &  1.28 &  1.32 &  1.69 \\
& EoH &  0.49 &  0.08 &  0.19 &  0.41 &  0.16 &  0.06 &  0.04 \\
& ReEvo &  0.36 &  0.35 &  0.22 &  0.45 &  0.33 &  0.28 &  0.58 \\
& (1+1)-EPS &  0.24 &  0.08 &  0.32 &  0.64 &  3.37 &  3.87 &  2.47 \\
    \bottomrule
    \end{tabular}
}
    \label{tab:top1_best}
\end{table}